\pdfoutput=1
\documentclass[]{simpleai}

\usepackage{graphicx}
\usepackage{booktabs}
\usepackage{tabularx}
\usepackage{array}
\usepackage{multirow}
\usepackage{makecell}
\usepackage{siunitx}
\usepackage{adjustbox}
\usepackage{wrapfig}
\usepackage{float}
\usepackage{caption}
\usepackage{amsmath,amsfonts,amssymb}
\usepackage{bm}
\usepackage{nicefrac}
\usepackage{enumitem}
\usepackage{xspace}
\usepackage{calc}
\usepackage{etoolbox}
\usepackage{pifont}
\usepackage{fancyvrb}

\setlist[itemize]{leftmargin=*}
\newcolumntype{C}{>{\centering\arraybackslash}X}
\crefname{figure}{Fig.}{Figs.}
\crefname{table}{Tab.}{Tabs.}
\crefname{section}{Sec.}{Secs.}

\title{
\textbf{HiFi-UMI: Learning Deployable Manipulation Policies from High-Fidelity UMI Data Alone}
}

\DeclareRobustCommand{\authorlink}[1]{%
  \begingroup\hypersetup{linkcolor=black}%
  \hyperref[sec:contributions]{#1}\endgroup}

\author[]{\authorlink{Simple AI}}

\abstract{
Learning deployable manipulation policies is bottlenecked by the scarcity of data that is both high-fidelity and scalable. Real-robot teleoperation is accurate but costly to scale; robot-free UMI capture scales readily, and current practice uses the resulting data mainly for pre-training, adding a small real-robot ``anchor'' at post-training time. We ask whether raising the \emph{fidelity} of robot-free UMI data---rather than shrinking the real-robot fraction---can remove that anchor. We present HiFi-UMI, a portable UMI data-production system that co-designs hardware and software for trajectory accuracy, inter-gripper relative pose, synchronization, and field of view: head-mounted \emph{offline} stereo-inertial SLAM, native rather than reconstructed relative pose, a shared microsecond GPIO trigger, and two wide-angle cameras per hand covering ${\sim}200^{\circ}$. It reaches $3$\,mm workspace-local end-effector accuracy with no external tracking infrastructure. Using this corpus, we demonstrate \emph{zero-robot} post-training: a policy post-trained \emph{solely} on HiFi-UMI demonstrations deploys directly on a real robot and matches in-domain teleoperation across three backbones spanning the vision-language-action (VLA) and world-action-model (WAM) families, with success-rate differences of $-2.5$, $+3.1$, and $-0.6$ percentage points on StarVLA-QwenPI, OpenPI-$\pi_{0.5}$, and LingBot-VA, respectively; the strongest policy reaches $85\%$ on a precision insertion task---even though the teleoperation baseline is collected in the evaluation scene and no HiFi-UMI trajectory is. Pre-training on $4{,}000$ hours from the same corpus lowers action error on ten unseen tasks by $41\%$ and, on StarVLA-QwenPI, raises real-robot success by a further $18.1$ percentage points. We open-source HiFi-UMI-2K---$2{,}000$ hours of microsecond-synchronized, ultra-wide-FoV demonstrations, each automatically reconstructed and validated through simulation replay---as a large-scale, high-fidelity resource for the robot-learning community.
}

\metadata[Website]{\texttt{https://cloud.simpleai.tech/simple-world-lab/hifi-umi/}}
\metadata[Dataset]{\texttt{https://huggingface.co/datasets/simple-world-lab/HiFi-UMI-2K}}

\begin{document}
\maketitle

\begin{figure}[!t]
  \centering
  \includegraphics[width=0.95\linewidth]{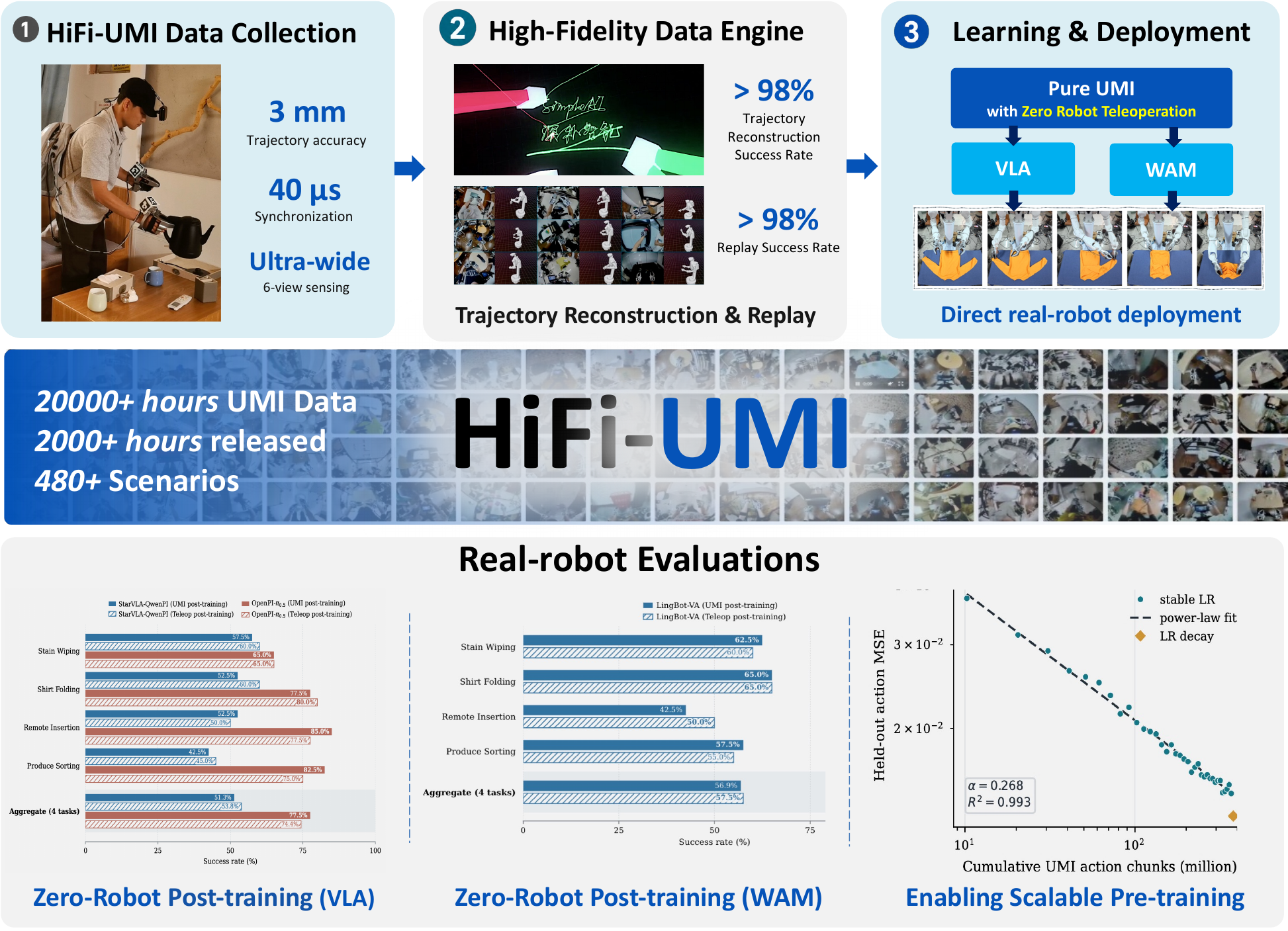}
  \caption{Overview of HiFi-UMI, a robot-free framework for scalable
  manipulation data collection, high-fidelity data processing, and direct real-robot
  policy deployment. HiFi-UMI Capture combines $3$\,mm trajectory accuracy,
  microsecond-level synchronization, and ultra-wide six-view sensing. The resulting
  data are reconstructed, replayed, quality-checked, annotated, and curated by the
  HiFi-UMI data engine, which has collected more than $20{,}000$ hours of data
  across over $480$ scenes, released the curated $2{,}000$-hour HiFi-UMI-2K
  subset, and achieved trajectory-reconstruction and replay validity rates above
  $98\%$ and $98\%$, respectively (${\approx}96\%$ cumulative). The curated corpus
  supports pre-training and post-training of both VLA and WAM policies without
  any teleoperated robot data for the target task, enabling
  direct deployment on real robots. HiFi-UMI-only post-training performs on par with in-domain teleoperation on
  all three backbones, differing by at most $3.1$ percentage points on the two VLA
  policies and by $0.6$ percentage points on the WAM policy. Moreover, the held-out action
  prediction error decreases consistently with increasing pre-training data, following
  a clear power-law scaling trend ($\alpha = 0.268$, $R^2 = 0.993$).}
  \label{fig:main}
\end{figure}

\section{Introduction}
\label{sec:intro}

Learning deployable manipulation policies has become bottlenecked not by model
capacity but by data. The dominant paradigm for acquiring deployable skills is
teleoperated demonstration on real robots~\citep{rt1,bridgev2,droid2024}, which
yields perfectly embodied, directly trainable trajectories but is expensive to
scale: it requires the target robot, a teleoperation rig, and a skilled operator
for every hour of data. Recent efforts show how steeply that cost scales: AgiBot
World assembled $2{,}976$ hours of manipulation data from $100$ dual-arm humanoid
robots in a purpose-built $4{,}000$\,m$^2$ facility~\citep{agibot}, and RoboMIND
assembled $305$ hours across four embodiments, each needing its own teleoperation
hardware built to match that arm~\citep{robomind}. As a result, the community has increasingly turned to cheaper, robot-free sources of manipulation data.

The Universal Manipulation Interface (UMI)~\citep{umi2024} is the most prominent of
these: a handheld, instrumented gripper that lets a human collect in-the-wild
demonstrations with no robot in the loop, at a fraction of the cost of
teleoperation. UMI and its derivatives have enabled large demonstration
corpora~\citep{fastumi} and even data-scaling studies of imitation
learning~\citep{datascaling}. Yet current handheld capture inherits a set of
fidelity deficiencies that limit how far such data can be trusted. Pose is
typically recovered by online visual(-inertial) SLAM, which drifts over long
horizons and fails under low texture or motion blur~\citep{umi2024,airexo2};
inter-gripper relative pose is reconstructed from cross-camera co-visibility rather
than measured natively, introducing error precisely on the coordinated tasks where it
matters most; sensor streams use software or wireless alignment rather than hardware
triggers; and one $155^{\circ}$ wrist fisheye per hand leaves blind spots and weak
depth cues~\citep{umi2024,airexo2}.

These limitations are not incidental---they are the reason that, throughout the
field, robot-free data is largely confined to a \emph{pre-training} role while
real-robot teleoperation is assumed necessary for the \emph{post-training} that
grounds a policy for deployment~\citep{groot,hrdt}. Even the most aggressive
recent attempts to minimize teleoperation preserve this division. The portable
VR-based systems ActiveUMI~\citep{activeumi} and XRZero-G0~\citep{xrzerog0}
reduce the real-robot fraction to a small share of a much larger robot-free
corpus, but neither eliminates it. RDT2~\citep{rdt2} obtains zero-shot
cross-embodiment transfer from robot-free data alone but mixes in real-robot
data once deployment-grade performance on a specific arm is required. It has
never been clear, however, whether post-training genuinely requires real-robot
data, or whether the fidelity of the robot-free data available so far has simply
not been high enough.

\begin{figure}[t]
  \centering
  \includegraphics[width=0.95\linewidth]{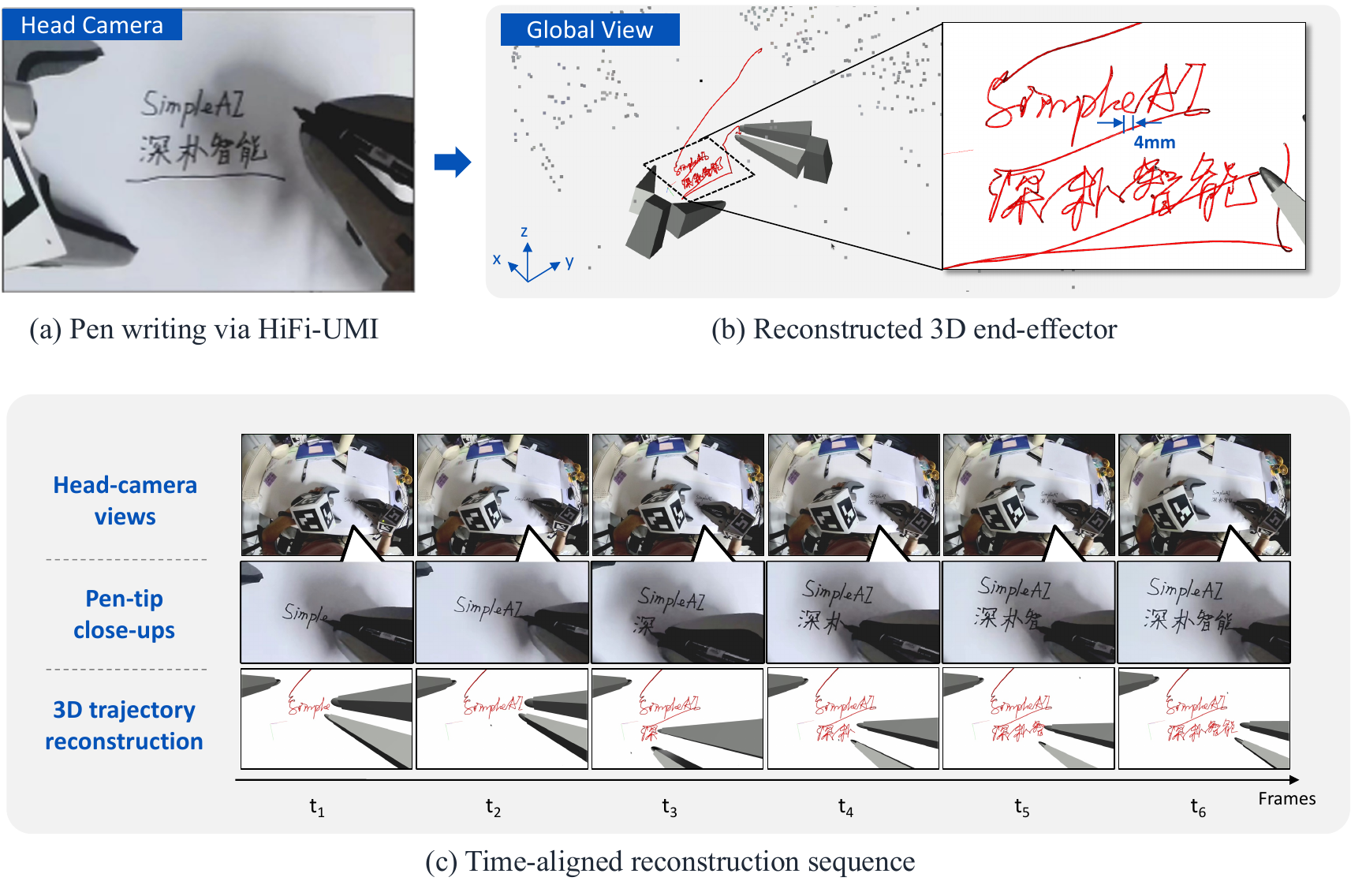}
  \caption{Qualitative visualization of 3D trajectory reconstruction in a handwriting task.
(a) First-person observation of a handwriting demonstration captured using HiFi-UMI.
(b) The reconstructed 3D end-effector trajectory, shown in a global view and a zoomed-in view of the written characters. The marked $4\,\mathrm{mm}$ width of the lowercase letter ``e'' provides a spatial reference for the scale of the reconstructed writing.
(c) Time-aligned visualization at six synchronized instants $t_1$--$t_6$, including the corresponding head-camera observations, pen-tip close-ups, and cumulative 3D trajectory reconstructions.}
  \label{fig:writing}
\end{figure}

We ask whether \emph{sufficiently high-fidelity} UMI data can break this
division. We present HiFi-UMI, a portable data-production system
designed end-to-end for fidelity, replayability, and automated curation
(\cref{fig:main}), and we use it to test a deliberately strong hypothesis---which
we call zero-robot post-training: that high-fidelity robot-free data
\emph{alone}, used for post-training, can yield manipulation policies that deploy
directly on a real robot without \emph{any} teleoperated real-robot data in
post-training. This
reframes the question away from mixing ratios. Where prior work approaches the
teleoperation baseline by shrinking the real-robot fraction to a small but
non-zero anchor, we ask whether raising the \emph{fidelity} of the robot-free data
itself---so that its trajectories, inter-gripper pose, and timing are as trustworthy as
teleoperation---can remove the need for that anchor altogether.
\Cref{fig:writing} makes this fidelity tangible: our offline reconstruction
preserves a handwriting trajectory well enough to render legible
millimeter-scale strokes.

HiFi-UMI attacks each of the deficiencies above. Pose fidelity comes from
head-mounted stereo \emph{offline} SLAM, which yields low-drift long-horizon
trajectories, together with natively accurate inter-gripper relative pose obtained
jointly with the world-frame pose of both hands. Sensing fidelity comes from
microsecond-level synchronization across all sensors via a single shared GPIO
hardware trigger, and from an ultra-wide field of view---two non-parallel stereo
cameras covering roughly $200^{\circ}$ horizontally and over $200^{\circ}$
vertically. Two further choices target the interaction itself and the cost of
producing usable data: a full-palm glove form factor that better preserves the
operator's natural force and contact than a trigger gripper, and real-time online
slicing with in-situ data-quality monitoring that catches capture anomalies during
collection rather than after it. Together these turn handheld capture into a
production-grade data engine that reconstructs, replays, quality-checks,
annotates, and curates every demonstration automatically---trajectory
reconstruction and replay validation each pass over $98\%$, a cumulative
$98\%\times98\%\approx96\%$---and it is this engine's output, and nothing else,
that we ask to carry a policy all the way to deployment. The same HiFi-UMI corpus
supplies both pre-training and post-training, yielding policies that run directly
on a real bimanual robot without teleoperated post-training data. These stages
correspond to the three panels of \cref{fig:main}.

The hypothesis holds across every comparison we run. On three backbones spanning
both the vision-language-action (VLA) and world-action-model (WAM) families,
HiFi-UMI-only post-training matches in-domain teleoperation: the differences are
$-2.5$, $+3.1$, and $-0.6$ percentage points, of both signs and each within the
sampling noise of our protocol. Parity holds under an asymmetry that favors the
baseline---the teleoperation data is collected in the evaluation scene and no
HiFi-UMI trajectory is---and the strongest policy reaches $85\%$ on a precision
insertion task. Because a claim of this kind is only as good as the evaluation
behind it, every comparison runs under a benchmark frozen before evaluation
begins, with test-instance construction separated from policy execution,
randomized policy order, and recorded termination reasons; the six conditions
receive $960$ real-robot rollouts in total. Using $4{,}000$ hours of the same
corpus for pre-training then lowers action error on ten unseen tasks by $41\%$
and, on StarVLA-QwenPI, raises real-robot success by a further $18.1$
percentage points---and the structure of that
transfer is itself informative, tracking coverage of interaction dynamics in the
pre-training mixture rather than whether a test object has been seen before. We
provide, to our knowledge, the first controlled demonstration that handheld
robot-free post-training, with no real-robot data at all, \emph{matches}
in-domain teleoperation on the same robot, replicated across three backbones.
We treat fidelity as the design principle
behind that result rather than a variable we isolate through controlled
degradation, and leave such an ablation---cleanly separating fidelity from sample
count and scene coverage---to future work.

In summary, the primary contributions of this work are:

\begin{itemize}
  \item A data-production system whose hardware--software co-design remedies the
  trajectory-accuracy, inter-gripper-pose, synchronization, and field-of-view
  deficiencies of prior handheld capture: head-mounted \emph{offline} stereo SLAM
  and a shared GPIO trigger give $3$\,mm end-effector accuracy and microsecond
  cross-sensor alignment with no external tracking infrastructure, and an automated
  engine reconstructs, replays, and validates every demonstration, retaining $96\%$
  of raw captures as robot-executable data.
  \item Evidence that HiFi-UMI alone \emph{suffices} for post-training: across
  three VLA and WAM backbones, UMI-only post-training matches in-domain
  teleoperation on the same robot ($-2.5$, $+3.1$, and $-0.6$ percentage points),
  with all gaps within sampling noise despite native sample-count differences.

  \item Pre-training on the same robot-free corpus raises both the data
  efficiency and the ceiling of downstream post-training: on StarVLA-QwenPI,
  $4{,}000$ hours cut offline action error on ten unseen tasks by $41\%$ and, at
  matched post-training data, raise real-robot success by $18.1$ percentage points, matching
  the scratch-initialized baseline with a quarter of the task data. Transfer
  depends more on whether pre-training covered a task's kind of physical
  interaction than on whether its objects have been seen.
  \item HiFi-UMI-2K, an open $2{,}000$-hour, microsecond-synchronized,
  replayable, ultra-wide-FoV subset, produced by the same pipeline for
  deployment-grade post-training without real-robot teleoperation.
\end{itemize}
\section{Related Work}
\label{sec:related}

\subsection{Manipulation Datasets}

Manipulation datasets now span a spectrum from fully grounded robot demonstrations
to scalable but weakly grounded human video. At the high-fidelity end, real-robot
teleoperation corpora remain the standard substrate for deployable policy learning:
BridgeData~V2~\citep{bridgev2} and RH20T~\citep{rh20t} established large-scale
multi-task and multimodal collection, DROID~\citep{droid2024} emphasized in-the-wild
diversity across scenes and operators, and Open~X-Embodiment~\citep{oxe2024}
aggregated heterogeneous robot datasets to study cross-embodiment transfer. More
recent efforts such as RoboMIND~\citep{robomind}, RoboMIND~2.0~\citep{robomind2},
and AgiBot World~\citep{agibot} push this regime toward larger trajectory scale
with stronger standardization, quality control, bimanual and mobile settings, and
richer sensory streams. At the opposite end, egocentric human-video corpora such as
Ego4D~\citep{ego4d} and Ego-Exo4D~\citep{egoexo4d} offer scale and naturalness but
lack executable robot actions, while EgoDex~\citep{egodex} narrows this gap with
manipulation-centric video and paired 3D hand-pose annotations.

Between these extremes, UMI-style data has emerged as a distinct middle tier of
\emph{robot-free} yet action-grounded supervision. UMI~\citep{umi2024} introduced
portable handheld grippers for collecting low-cost, information-rich demonstrations
without a robot body, enabling direct transfer to hardware-agnostic policies.
FastUMI-100K~\citep{fastumi100k} and the $10{,}000$-hour corpus of
RDT2~\citep{rdt2} show that this recipe scales to corpora rivaling the largest
teleoperation efforts; we defer the corresponding devices to
\cref{sec:related:devices}. This hierarchy is naturally viewed as a \emph{data pyramid}~\citep{groot}:
web and human video provide scale, teleoperation provides embodiment-specific
grounding, and UMI-style demonstrations occupy the middle by preserving actionable
wrist-view geometry and relative end-effector motion without robot-specific
collection. Our work targets this middle tier directly: by increasing the
fidelity, synchronization, and retargetability of UMI data, we test whether
robot-free handheld demonstrations can serve not only as scalable pre-training
data but also as deployment-relevant supervision.

\subsection{UMI and Portable Data-Collection Devices}
\label{sec:related:devices}

The Universal Manipulation Interface~\citep{umi2024} introduced a handheld
instrumented gripper that recovers global-scale end-effector trajectories via
ORB-SLAM3 and an IMU, using a single wrist-mounted $155^{\circ}$ fisheye camera
per gripper and side mirrors for implicit depth. It is the foundation for a
growing family of portable capture devices. \Cref{tab:umi-compare} compares them
along the axes that matter here: pose acquisition and accuracy, cross-sensor
synchronization, sensing coverage, whether inter-gripper relative pose is
measured natively or reconstructed, gripper form factor, and portability.
FastUMI~\citep{fastumi} swaps bespoke SLAM for an off-the-shelf tracker,
improving robustness and scaling collection. Its successor FastUMI~Pro, the
capture platform used by VISTA~\citep{vista}, fuses an external lighthouse tracker with onboard visual-inertial SLAM,
reaching millimeter-level pose at the cost of fixed infrastructure.
DexCap~\citep{dexcap} and DexUMI~\citep{dexumi} extend capture to dexterous
hands via mocap gloves and wearable exoskeletons, and DexWild~\citep{dexwild}
scales in-the-wild human-hand demonstrations for dexterous policies.
ARCap~\citep{arcap} adds in-headset augmented-reality feedback so demonstrations
remain kinematically valid, and exoskeleton systems such as
AirExo~\citep{airexo,airexo2} pursue whole-arm capture without a robot. A
parallel line pairs head cameras with hand capture---EgoMimic~\citep{egomimic}
and H-RDT~\citep{hrdt}---to obtain action signals.

Despite this progress, fidelity limitations remain pervasive across these
devices, and the methods that consume robot-free data still lean on paired robot
demonstrations. AirExo-2~\citep{airexo2} explicitly attributes the shortcomings
of handheld UMI-style devices to two causes: reliance on visual SLAM for pose
estimation, which yields action inaccuracies, and a limited camera field of
view, which hinders depth perception. The original UMI~\citep{umi2024} likewise
notes SLAM and scale-ambiguity failures, latency discrepancies between
collection and inference, and constrained single-camera coverage. Crucially,
methods that do leverage human or egocentric data for learning still rely on
\emph{paired} robot data: EgoMimic~\citep{egomimic} co-trains with teleoperated
demonstrations, and H-RDT~\citep{hrdt} fine-tunes on robot data after human
pre-training. In this line, robot-free data is not asked to ground a deployable
policy on its own.

A recent line of work replaces on-device handheld SLAM with tracking
infrastructure outside the gripper, whether a headset worn by the operator or
base stations placed in the room. ActiveUMI~\citep{activeumi} rigidly mounts a
copy of the robot's gripper onto a VR controller and records the operator's head
motion and egocentric attention. XRZero-G0~\citep{xrzerog0} pairs headset
tracking with dual purpose-built grippers and a closed-loop quality-inspection
pipeline to build large robot-free datasets. These systems improve tracking
robustness and, like ours, recover the inter-gripper relative pose natively,
since the headset observes both controllers together. Their remaining design
choices differ from ours in ways that matter for fidelity. Pose comes from
\emph{online} headset tracking rather than \emph{offline} SLAM optimization;
sensor streams are aligned by software spatiotemporal matching rather than a
hardware trigger; and coverage comes from a small number of discrete camera
views rather than ultra-wide stereo optics. They reduce the real-robot fraction
to a small share of a much larger robot-free corpus, but neither eliminates it.
RDT2~\citep{rdt2} instead scales redesigned UMI hardware to over 10{,}000 hours
and attains zero-shot cross-embodiment transfer on simple open-vocabulary tasks
with no real-robot data. Only when deployment-grade performance on a specific
arm is required does it mix a small amount of real-robot data into an optional
post-training variant. That hardware, however, tracks the end-effector with
external infrared base stations rather than onboard SLAM, so every collection
site must first be instrumented. Closest to our setting, VISTA~\citep{vista}
post-trains a bimanual policy on curated handheld data alone and deploys it on
real robots, establishing that robot-free post-training can work. Because all of
its baselines are trained on that same handheld corpus, however, the comparison
isolates model and curation design rather than the data source, and the question
of how robot-free supervision stands against \emph{teleoperation} is left open.
We share this emphasis on post-collection validation, but also build fidelity
into the capture device itself, so that most trajectories pass validation rather
than being screened out.

Our system addresses these fidelity limitations through offline stereo SLAM, a
shared GPIO hardware trigger, and non-parallel cameras for ultra-wide coverage.
No prior handheld system places robot-free post-training against teleoperation
on the same robot~\citep{activeumi,xrzerog0,rdt2,vista}. We instead hold the
backbone, the recipe, and the deployment stack fixed and change only whether the
task-specific demonstrations come from HiFi-UMI or from teleoperation on the
evaluation robot.

\subsection{Manipulation Foundation Models}

Recent manipulation foundation models fall into two families of low-level control
backbone, separated by whether action generation is coupled to a prediction of the
future. VLA policies are purely reactive, mapping the current observation and a
language instruction directly to actions. World-action models (WAMs) add a
predictive component, and that coupling takes two forms. Some predict future
observations only as an auxiliary training signal and discard the predictor at test
time. Others generate a future at every inference step and decode the action from
it, so that the quality of the imagined future directly gates control.

Within the VLA family, RT-2~\citep{rt2} established the tokenized-action recipe,
and OpenVLA~\citep{openvla} and Octo~\citep{octo} scaled open generalist policies
over heterogeneous robot data. Recent systems favor continuous action heads and
stronger post-training recipes: $\pi_0$~\citep{pi0} and $\pi_{0.5}$~\citep{pi05}
pair a pretrained VLM with a flow-matching action expert, an architecture now
widely reused, while GR00T~N1~\citep{groot}, GR-3~\citep{gr3}, Gemini
Robotics~\citep{geminirobotics}, and SmolVLA~\citep{smolvla} trade model scale
against inference cost and deployability. A separate line treats the action
interface itself as the design variable, with FAST~\citep{fast} on action
tokenization and OpenVLA-OFT~\citep{openvlaoft} on chunked decoding and
fine-tuning; our own action representation inherits these choices
(\cref{sec:setup}). LingBot-VLA~\citep{lingbotvla}, Qwen-VLA~\citep{qwenvla}, and
Qwen-RobotManip~\citep{qwenrobotmanip} pursue large-scale aligned pre-training and
unified embodied interfaces.

Within the WAM family, early systems explored the idea before the terminology
settled. GR-1~\citep{gr1} and GR-2~\citep{gr2} jointly predict future images and
actions after video-generative pre-training, UniPi~\citep{unipi} casts policy
learning as text-guided video generation, and VPP~\citep{vpp} conditions control on
representations drawn from a video diffusion model. Recent WAMs make the coupling
explicit in the policy backbone: DreamZero~\citep{dreamzero} builds a real-time
closed-loop policy on a pretrained video-diffusion backbone, and
WorldVLA~\citep{worldvla} models image and action generation jointly in an
autoregressive framework. LingBot-VA~\citep{lingbotva} learns frame prediction and
policy execution through causal world modeling, decoding each action chunk from the
future latents it has just generated.

These trends reinforce the value of scalable data. Yet robot-free sources such as
human video, egocentric video, and UMI-style demonstrations are still rarely used as
post-training supervision, a gap commonly attributed to limited trajectory fidelity,
synchronization, retargetability, and geometric consistency. StarVLA~\citep{starvla}
and its QwenPI-style implementations provide an open, modular ecosystem in which
that attribution can be tested directly.

We therefore select three backbones that differ along three axes. StarVLA-QwenPI is
a modular open implementation, so we control its initialization and can add a
large-scale pre-training arm. OpenPI-$\pi_{0.5}$ is a strong publicly released
checkpoint that we did not build. LingBot-VA derives actions from an imagined future
rather than from the current observation alone. What makes them comparable is not
their architecture but their interface: each consumes the same supervision, however
differently it tensorizes it. Changing the source of that supervision is therefore
one intervention applied three times, and agreement among the three is evidence
about the data rather than about any single architecture.
\section{Data Collection and Processing Pipeline}
\label{sec:pipeline}

The core hypothesis of this work---that robot-free demonstrations can, on their
own, ground a deployable policy---stands or falls on the fidelity of the data.
If trajectories drift over long horizons, if sensor streams are misaligned in time, or if the gripper's
view is too narrow to resolve contact, then no amount of scale recovers a
deployment-grade action signal~\citep{groot}, and a real-robot anchor becomes
unavoidable. We
therefore treat data production as a \emph{system-design} problem and co-design
hardware and software so that fidelity is enforced at the source rather than
repaired after the fact. This section describes the resulting system: a wearable
capture device whose four subsystems each target a specific fidelity axis
(\cref{sec:pipeline:hardware}); an explicit, two-level definition of what
constitutes usable data (\cref{sec:pipeline:quality}); a six-stage processing
flywheel that turns raw captures into training-ready episodes
(\cref{sec:pipeline:processing}); and the end-to-end fidelity the system
delivers (\cref{sec:pipeline:processed}).

\subsection{HiFi-UMI Capture Device}
\label{sec:pipeline:hardware}

\begin{figure}[h]
  \centering
  \includegraphics[width=0.95\linewidth]{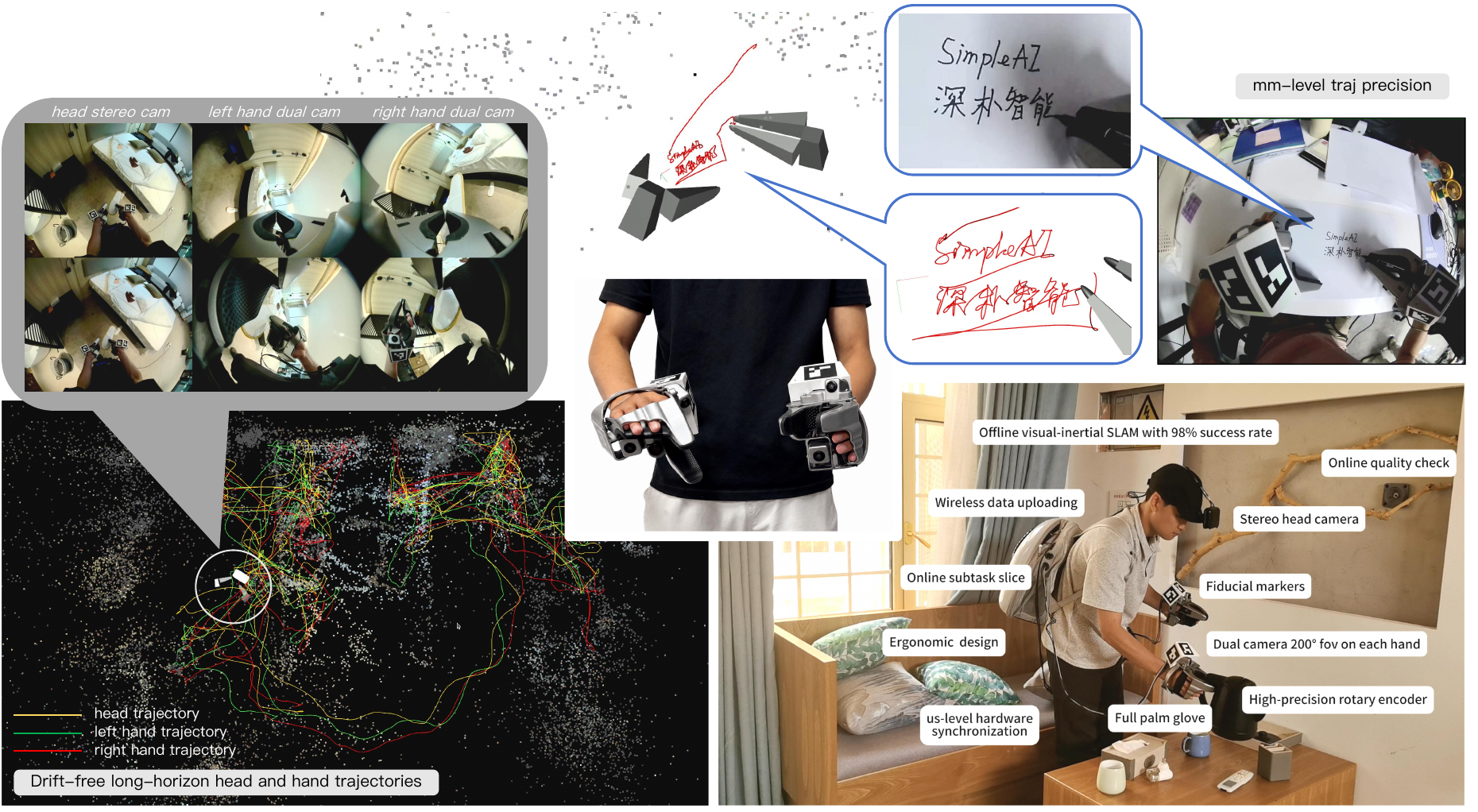}
  \caption{Overview of the HiFi-UMI capture device. A head-mounted stereo camera pair with an integrated IMU enables offline stereo-inertial SLAM; per-hand marker cubes
  are localized in the same head-camera frame; each hand carries two
  non-parallel wide-angle fisheye cameras (top and bottom) for ultra-wide
  coverage; a full-palm glove gripper preserves natural contact; and a shared
  GPIO trigger synchronizes all sensors.}
  \label{fig:hardware-overview}
\end{figure}

The HiFi-UMI capture device (\cref{fig:hardware-overview}) is organized around four
requirements that our design treats as jointly sufficient for
deployment-grade robot-free data: (i)~accurate and scalable pose acquisition,
(ii)~a manipulation-oriented gripper morphology, (iii)~wide-coverage multimodal
sensing, and (iv)~online quality control. We address each in turn, and
summarize how our design compares to representative robot-free systems in
\cref{tab:umi-compare}.

\subsubsection{Pose Acquisition and Accuracy}

The core tension in UMI-style capture~\citep{umi2024} is to reconcile trajectory
accuracy,
tracking robustness, deployment flexibility, and hardware cost. Existing
approaches sit at different points of this trade-off: wrist-camera visual-inertial
odometry (VIO)~\citep{vinsmono}, VR headset-and-controller tracking, base-station
tracking, and motion capture. Base-station and motion-capture systems deliver high
accuracy but require instrumented environments, which precludes scalable in-the-wild
collection. VR-headset solutions inherit mature commercial tracking stacks, at the
cost of higher hardware expense and greater system complexity. Wrist-camera VIO is
cheaper and lighter than either, yet its view is routinely occluded by the hand or
the manipulated object, leaving trajectories vulnerable to tracking failure and
accumulated drift.

We instead adopt a wearable scheme built on \emph{offline} stereo-inertial
SLAM~\citep{orbslam3} together with fiducial-marker localization~\citep{apriltag}. Rather than tracking each wrist
independently, a head-mounted stereo rig estimates the global camera trajectory,
and each hand is localized \emph{relative to the head} via a rigidly attached
marker cube observed by the same head cameras. Composing the global head
trajectory with the two relative hand-to-head poses yields globally consistent
trajectories for both hands. This design is motivated by a simple observation:
a head-mounted viewpoint is far more stable than a wrist-mounted one, which is
routinely corrupted by nearby moving objects, self-occlusion, and rapid hand
motion during manipulation~\citep{airexo2}. In our experiments the scheme attains $3$\,mm
end-effector accuracy---comparable to VR-controller tracking---while remaining
lighter and lower-cost than VR-headset-based rigs.

The head-relative formulation is especially advantageous for bimanual
manipulation. Because both marker cubes are observed in a single head-camera
frame, the inter-gripper relative pose is measured \emph{natively} and inherits
the same accuracy as the per-gripper pose, rather than being reconstructed post
hoc from cross-camera co-visibility as in prior handheld rigs. Moreover, because
head motion is typically far smaller than hand motion, head-relative tracking
substantially reduces accumulated drift.

\subsubsection{Gripper Morphology}

Existing UMI grippers fall broadly into trigger-based and finger-sleeve-based
designs. Trigger devices emulate parallel-jaw commands directly, but give the
operator weak tactile correspondence with the manipulated object. Finger-sleeve
designs are mechanically simpler and preserve direct contact sensation, enabling
more natural and dexterous manipulation. Many, however, adopt a narrow, elongated
geometry: well suited to fine manipulation, but limiting on larger or heavier
objects.

To broaden task coverage without sacrificing contact fidelity, we design an
asymmetric two-finger, glove-like \emph{full-palm} gripper inspired by the human
hand. The two fingers correspond respectively to the thumb and the opposing four
fingers, and are deliberately asymmetric in shape and width: the narrower
fingertip region supports precise, small-object manipulation, while the wider
proximal region provides a larger contact patch and firmer support for heavy
objects. This morphology preserves the operator's natural force distribution and
contact geometry more faithfully than a trigger interface.

\subsubsection{Cameras and Sensors}

Each hand carries two non-parallel fisheye cameras, yielding about $200^{\circ}$
of horizontal and vertical coverage. This ultra-wide view reduces occlusion and
improves observability around the gripper.
Together with the stereo head cameras, the complete device integrates six
cameras. It further carries IMUs on the head and both hands---for pose estimation
and motion-state monitoring---and high-precision encoders on the grippers to
measure opening angle. Critically, every sensor is driven by a single, unified
GPIO external trigger~\citep{versavis}, providing microsecond-level temporal
synchronization across all cameras, IMUs, and encoders. This hardware-level synchronization
replaces the software or wireless alignment used by prior systems and removes a
key source of action-label noise.

\subsubsection{Online Interaction and Quality Control}

To suppress low-quality data at its source, the device performs quality
monitoring \emph{during} recording. It detects common failure modes---including
underexposure, motion blur, excessively fast operator motion, and hands leaving
the head cameras' field of view---and issues real-time voice feedback so the
operator can correct the capture in situ rather than discovering the problem in
post-processing.
Online temporal slicing lets operators mark task and subtask boundaries during
collection, reducing later segmentation effort. Alongside online monitoring, it
improves throughput and reliability.

\begin{table*}[t]
  \centering
  \caption{
  Comparison of representative robot-free data-collection systems along key design
  and data-fidelity axes.
  \textbf{Pose acquisition}: how 6-DoF pose is estimated---on-device
  visual(-inertial) odometry/SLAM (VIO / VI-SLAM), headset inside-out tracking
  (VR inside-out), or external base-station tracking.
  \textbf{Pos.\ err.}: reported end-effector positional error (mm), taken from each
  system's own publication and rounded; because each is measured against a different
  reference and motion profile, these values should be read as order-of-magnitude
  rather than compared directly. The HiFi-UMI value is measured in this work
  (\cref{tab:fidelity}).
  \textbf{Sync.}: cross-sensor time-alignment, reported as its typical latency scale
  and mechanism; only HiFi-UMI reaches microsecond-level alignment via a hardware
  GPIO trigger, whereas prior systems, where reported, use millisecond-level
  software timestamp alignment.
  \textbf{Views\,/\,FoV}: number of camera views and peak field-of-view coverage.
  \textbf{Rel.\ pose}: how the inter-gripper relative pose is obtained---measured
  natively when both ends are observed together in one operator-mounted frame
  (\emph{native}), reconstructed post hoc from cross-camera co-visibility
  (\emph{reconstructed}), or differenced from two poses tracked against external
  infrastructure (\emph{external}).
  \textbf{Gripper}: end-effector actuation form factor (e.g.\ handheld trigger,
  finger-sleeve, or full-palm glove).
  \textbf{Port.}: portability---untethered, in-the-wild capture with no instrumented
  infrastructure (High) vs.\ dependence on fixed base stations or motion-capture rigs
  (Low).
  ``--'' marks a property not reported by the cited source or not applicable to
  single-arm capture.
  Together, these choices yield HiFi-UMI's main advantages over prior systems:
  millimeter-level end-effector accuracy (${\sim}3$\,mm) obtained from head-mounted
  offline stereo-inertial SLAM without external tracking infrastructure; the
  tightest, microsecond-level synchronization via a GPIO hardware trigger; the widest
  sensing coverage at both hands; and greater ease of use---fully portable with no external base stations, and
  operated through a full-palm glove rather than a trigger for more natural
  manipulation.
  }
  \label{tab:umi-compare}
  \scriptsize
  \setlength{\tabcolsep}{3pt}
  \renewcommand{\arraystretch}{1.05}
  \begin{tabularx}{\linewidth}{@{}l >{\raggedright\arraybackslash}X c c c c >{\raggedright\arraybackslash}X c@{}}
    \toprule
    System
      & Pose acquisition
      & \makecell{Pos.\\err.\,(mm)}
      & Sync.
      & \makecell{Views\,/\\FoV}
      & \makecell{Rel.\\pose}
      & Gripper
      & Port. \\
    \midrule
    UMI~\citep{umi2024}
      & wrist VI-SLAM             & ${\sim}6$   & ms (software) & 2 / $155^{\circ}$ & reconstructed & trigger                   & High \\
    FastUMI~\citep{fastumi}
      & dedicated VI module (T265) & ${\sim}10$ & ms (software) & 1 / $155^{\circ}$ & --            & trigger                   & High \\
    DAS fingers~\citep{dasgripper}
      & wrist VIO                 & --          & --            & 2 / $150^{\circ}$ & reconstructed & finger-sleeve             & High \\
    ActiveUMI~\citep{activeumi}
      & VR inside-out             & ${\sim}4$   & ms (software) & 3 / --            & native        & trigger                   & High \\
    TacUMI~\citep{tacumi}
      & base station              & --          & --            & 1 / --            & external      & trigger                   & Low  \\
    RDT2~\citep{rdt2}
      & base station              & --          & --            & 2 / --            & external      & trigger                   & Low  \\
    FastUMI Pro~\citep{vista}
      & base station + wrist VIO  & ${\sim}3$   & --            & 2 / $180^{\circ}$ & external      & trigger                   & Low  \\
    XRZero-G0~\citep{xrzerog0}
      & VR inside-out             & ${\sim}4$   & --            & 3 / --            & native        & trigger and finger-sleeve & High \\
    \midrule
    \textbf{HiFi-UMI (Ours)}
      & \textbf{head stereo-inertial SLAM}
      & $\mathbf{{\sim}3}$
      & \textbf{$\mu$s (GPIO)}
      & \textbf{6 / $200^{\circ}$}
      & \textbf{native}
      & \textbf{full-palm glove}
      & \textbf{High} \\
    \bottomrule
  \end{tabularx}
\end{table*}

\subsection{Data-Quality Criteria}
\label{sec:pipeline:quality}

Data quality is enforced as a set of hard constraints on sensors, trajectories,
and annotations---the technical requirements a capture must meet before it can
serve as an action label---mirroring the curation and quality-control protocols
of large-scale robot datasets~\citep{droid2024}. Beyond these constraints, the
content-level properties of the corpus---annotation correctness, task and scene
composition, and coverage of rare behaviors---are governed by the annotation,
verification, and export stages of the processing pipeline
(\cref{sec:pipeline:processing}) rather than by a separate admission test.

\noindent\textbf{Sensors.} Camera field of view, layout, and viewing direction
must satisfy the system specification; multi-sensor timestamps must be accurately
synchronized; images must be properly exposed, free of severe motion blur, and
free of unexpected occlusion; and all other sensor readings must be valid and
accurate.

\noindent\textbf{Trajectories.} The coordinate frames of every 6-DoF trajectory
must be consistently and explicitly defined; the reconstructed trajectories must
faithfully reflect the human demonstration, meet the required precision, and be
executable under robot replay; and the gripper width or opening angle must be
measured accurately.

\noindent\textbf{Annotations.} Language annotations must be consistent with the
visual content, and the temporal boundaries of subtask segments must meet the
required timing accuracy.

These constraints are secured primarily by the device design and the processing
pipeline. The current usable-data ratio is approximately $96\%$, which is itself
cumulative: it is the product of the two gates applied in series by the
pipeline---trajectory reconstruction and whole-body-control replay
validation---each of which passes approximately $98\%$ of the captures that
reach it (\cref{sec:pipeline:processing}).

\subsection{HiFi-UMI Data Processing Pipeline}
\label{sec:pipeline:processing}

\begin{figure}[h]
  \centering
  \includegraphics[width=\linewidth]{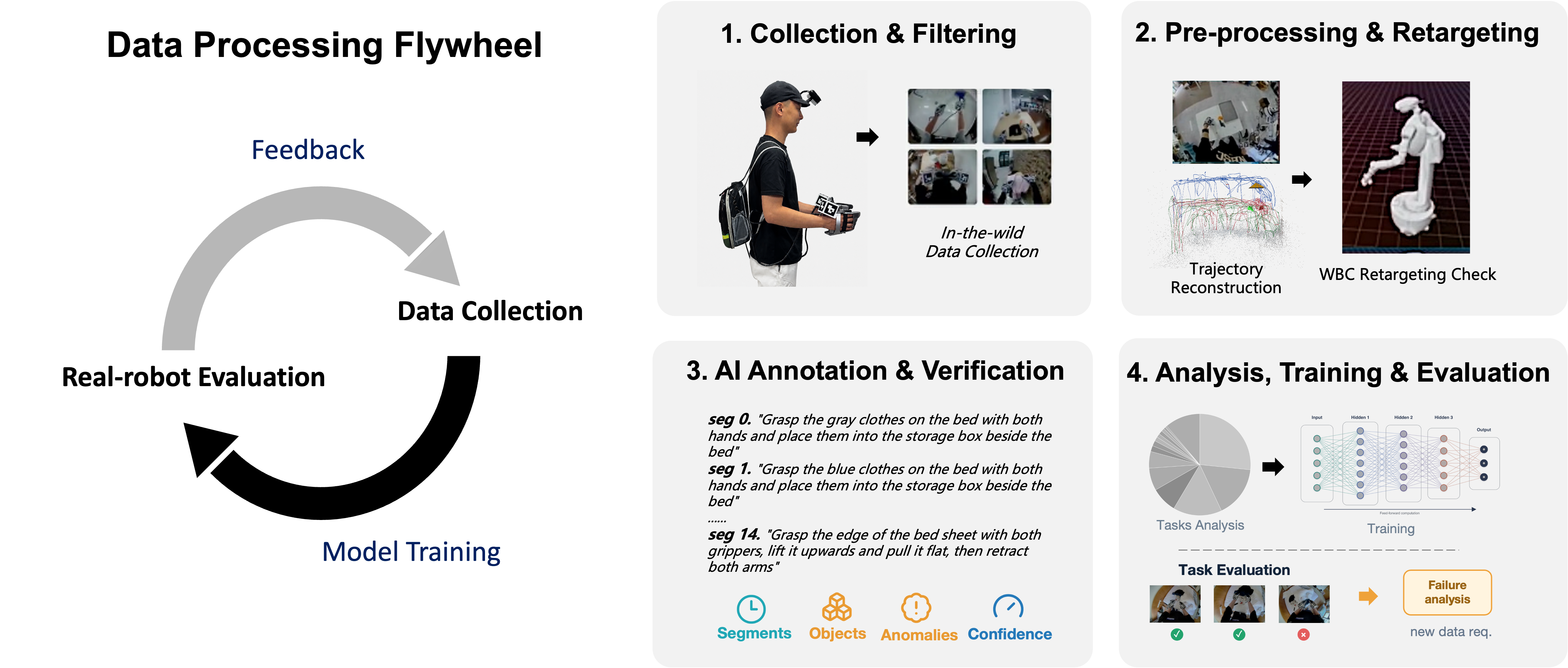}
  \caption{The HiFi-UMI data flywheel. Raw captures pass through six
  stages---collection and upload, trajectory reconstruction and automatic
  cleaning, simulation retargeting, AI-assisted annotation, human verification,
  and analysis and export---each of which enriches the data with metadata and
  removes or flags invalid samples.}
  \label{fig:system-overview}
\end{figure}

The HiFi-UMI pipeline (\cref{fig:system-overview}) comprises six stages: data collection
and upload, trajectory reconstruction and automatic cleaning, simulation
retargeting, AI-assisted annotation, human verification, and data analysis and
export. Each stage both filters invalid data and augments valid data with
structured metadata, so that quality control is distributed across the pipeline
rather than deferred to a single expensive review step.

\subsubsection{Data Collection and Upload}

Four mechanisms operate at capture time. Multi-sensor synchronization is
guaranteed in hardware via the shared GPIO trigger; on-device processing raises
online warnings for corrupted or low-quality segments, removing invalid samples at
the source; a human-in-the-loop interface lets operators mark subtask boundaries
during collection, reducing downstream segmentation cost; and the device streams
captures to the cloud over Wi-Fi in real time, so collection and upload proceed
concurrently.

\subsubsection{Trajectory Reconstruction and Automatic Cleaning}

\begin{figure}[h]
  \centering
  \includegraphics[width=0.92\linewidth]{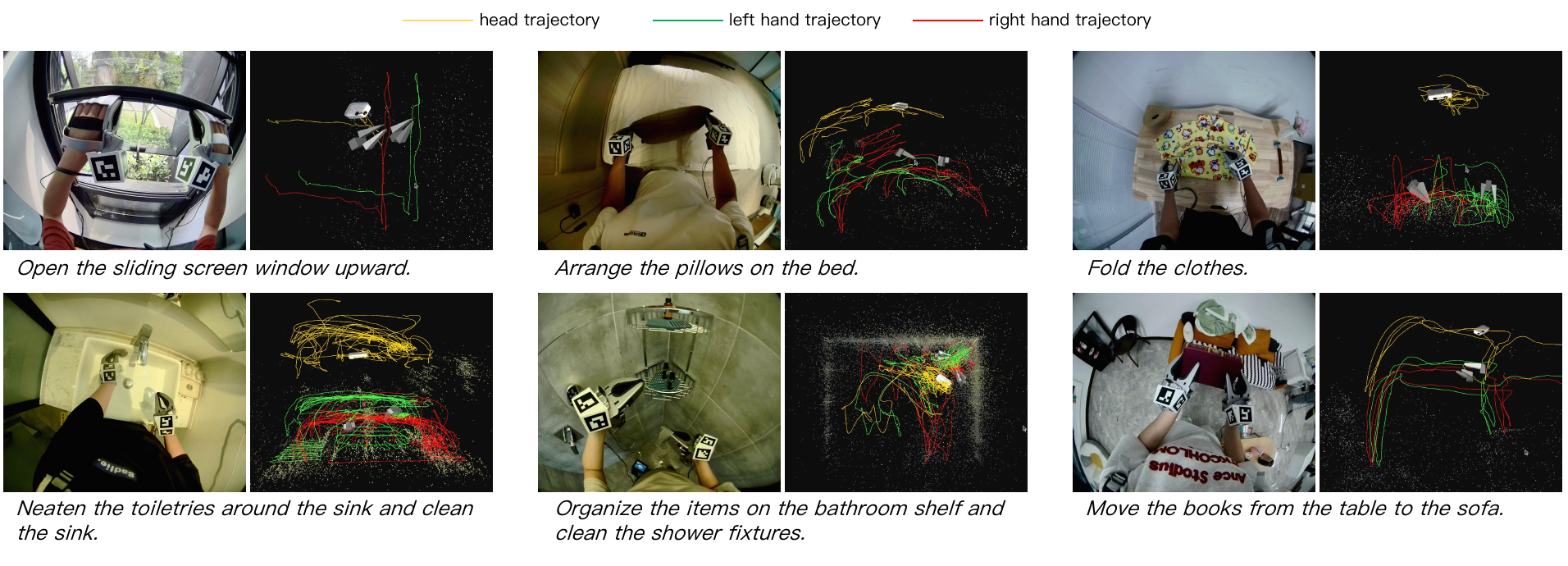}
  \caption{Representative HiFi-UMI tasks: head-camera views (left) and 3D
  end-effector trajectories (right).}
  \label{fig:data-trajs}
\end{figure}

We reconstruct the head trajectory with offline stereo-inertial SLAM and recover
both hand trajectories by detecting their fiducial markers in the head cameras
(\cref{fig:data-trajs}).
Operating \emph{offline} lets the optimizer exploit future as well as past
observations. Because manipulation continually alters the scene---violating the
static-world assumption underlying standard loop closure~\citep{dynaslam}---we
deliberately forgo global loop closure and instead impose a local-consistency constraint over a
dynamic sliding window, which bounds global drift to the centimeter level over
long horizons while preserving the millimeter-level \emph{local} accuracy reported
above. As SLAM optimization exhibits occasional stochastic failures, trajectories
flagged as abnormal are automatically recomputed. A subsequent automatic-cleaning
pass detects and annotates residual issues such as abnormal SLAM estimates,
inconsistent trajectories, or other anomalous data patterns. This stage
reconstructs $98\%$ of trajectories; the remaining failures are automatically
detected and removed.

\subsubsection{Simulation Retargeting}

Whether a trajectory can be replayed on the target robot is a prerequisite for
its use in policy training. We develop a whole-body motion-control algorithm~\citep{umionlegs} for
the target embodiment and validate every reconstructed trajectory by replaying it
in simulation, discarding trajectories that are kinematically or dynamically
infeasible. In our experiments this replay validation succeeds for $98\%$ of
reconstructed trajectories. Because reconstruction and replay validation are
applied in series, the cumulative basic-validity yield is
$98\%\times98\%\approx96\%$ of raw captures, the figure reported in
\cref{sec:pipeline:quality}.

\subsubsection{AI-Assisted Annotation}

An annotation model supplements subtask segmentation and generates draft
labels~\citep{robovqa}.
Exploiting the multi-view captures unique to our device, the model reasons jointly
over the head-mounted and hand-centric views to infer task progress, object
interactions, and action boundaries; this multi-view design resolves ambiguity in
cases where the object of interest is occluded in one view but visible in another.
The stage emits structured metadata---task- and subtask-level language
descriptions, temporal segment boundaries, manipulated objects, and candidate
abnormal events---giving each demonstration an initial structured representation
that substantially reduces later manual effort. Each label carries a confidence
or uncertainty score, so that low-confidence samples can be routed preferentially
to human review.

\subsubsection{Human Verification}

Human annotators perform sampling-based inspection and final verification. Rather
than reviewing all raw data from scratch, they concentrate on samples flagged by
automatic quality checks, low-confidence AI annotations, or distribution-level
analysis, which keeps manual effort low without sacrificing reliability. During
verification they confirm that language descriptions match the visual content,
that subtask boundaries are temporally accurate, and that each demonstration
satisfies its intended task definition; they correct or supplement AI labels---for
example, adjusting temporal boundaries, refining coarse descriptions, adding
missing object information, or marking samples for removal or down-weighting.

Verification outcomes are stored as structured quality-control metadata, so that
every sample can be traced to its annotation source, review status, rejection
reason, and correction history. Such traceability is essential at scale: it
enables downstream analysis of failure modes, annotator consistency, and
data-quality trends across tasks, scenes, devices, and collection batches.

\subsubsection{Data Analysis and Export}

Finally, the curated data is analyzed statistically before export. The analysis
summarizes the dataset along multiple dimensions---task category, scene type,
object category, action pattern, trajectory quality, annotation quality, and
replay success---yielding a quantitative view of the distribution that surfaces
over-represented, under-represented, or otherwise imbalanced subsets. Guided by these
statistics, training sets are assembled by explicitly balancing task attributes
against the requirements of the target model: the export process can control the
proportions of scenes, tasks, objects, action types, and recovery behaviors,
down-weight redundant subsets, or up-weight rare but important behaviors. We
deliberately collect rare failure-and-recovery episodes to provide supervision
for robust closed-loop execution.

Export converts the curated data into training-ready formats. Beyond raw
observations and action trajectories, each exported episode carries synchronized
multi-view video, calibrated trajectories, gripper states, language annotations,
subtask boundaries, and quality-control metadata, so that a single data lake can
serve heterogeneous training configurations---task-level imitation learning,
subtask-conditioned policy learning, VLA training, and offline
evaluation. In effect, this analysis-and-export stage closes the loop between
collection and training: instead of treating all demonstrations as equally useful,
the system constructs datasets through explicit filtering, balancing, and
versioned export, improving data utilization and making downstream model
performance traceable to specific data choices.

\subsection{Processed-Data Quality}
\label{sec:pipeline:processed}

\cref{tab:fidelity} reports the end-to-end fidelity of the processed data. The
pipeline delivers $3$\,mm end-effector accuracy, cross-sensor timing offsets
below $40\,\mu\mathrm{s}$, fewer than two dropped frames per hour of capture, a
$98\%$ trajectory-reconstruction success rate, and gripper-state error below
$0.1^{\circ}$.
Within a typical manipulation workspace---on the order of $2$\,m of accumulated
head-trajectory length---the recovered end effector attains a mean translational
error of $3$\,mm against base-station tracking ground truth (used only for this
accuracy evaluation, not for routine capture). This is the regime our target
manipulation scenarios operate in: they depend on locally consistent trajectories
and accurate relative hand poses within the workspace.
Taken together, these numbers indicate that the trajectory accuracy, timing, and
gripper reconstruction of HiFi-UMI captures are on par with what real-robot
teleoperation provides---the property our central hypothesis requires.

Fidelity must, however, survive into the policy's action space. We represent
actions in a robot-centric end-effector frame, predicting \emph{relative} pose
increments together with an absolute gripper opening. We detail this action
representation in \cref{sec:setup}.

\begin{table}[t!]
  \centering
  \caption{End-to-end fidelity of the processed data.}
  \begin{tabularx}{\linewidth}{@{}lXl@{}}
    \toprule
    Metric & Description & Value \\
    \midrule
    Pose accuracy        & Local end-effector error (${\sim}2$\,m workspace)   & $3$\,mm \\
    Synchronization      & Cross-sensor timing offset      & $<40\,\mu\mathrm{s}$ \\
    Frame-drop rate      & Dropped frames (6 cameras @ 25\,fps) & $<1$ per $270{,}000$ frames \\
    Reconstruction        & Trajectories passing SLAM reconstruction & $98\%$ \\
    Gripper-state error  & Opening-angle error             & $<0.1^{\circ}$ \\
    \bottomrule
  \end{tabularx}
  \label{tab:fidelity}
\end{table}

\section{Dataset and Release}
\label{sec:dataset}

The dataset is the direct product of the pipeline of \cref{sec:pipeline}. To
date it comprises over $20{,}000$ hours of processed manipulation data spanning
more than $4.32$~million episodes across $480{+}$ scenes. Every episode is
captured with the six-camera configuration of \cref{sec:pipeline:hardware}---stereo
head views plus two ultra-wide fisheye views per hand---and is exported with
synchronized multi-view video, calibrated bimanual trajectories, gripper states,
language annotations, and subtask boundaries. \cref{tab:dataset-stats}
summarizes the headline statistics.

\begin{figure}[h]
  \centering
  \includegraphics[width=0.92\linewidth]{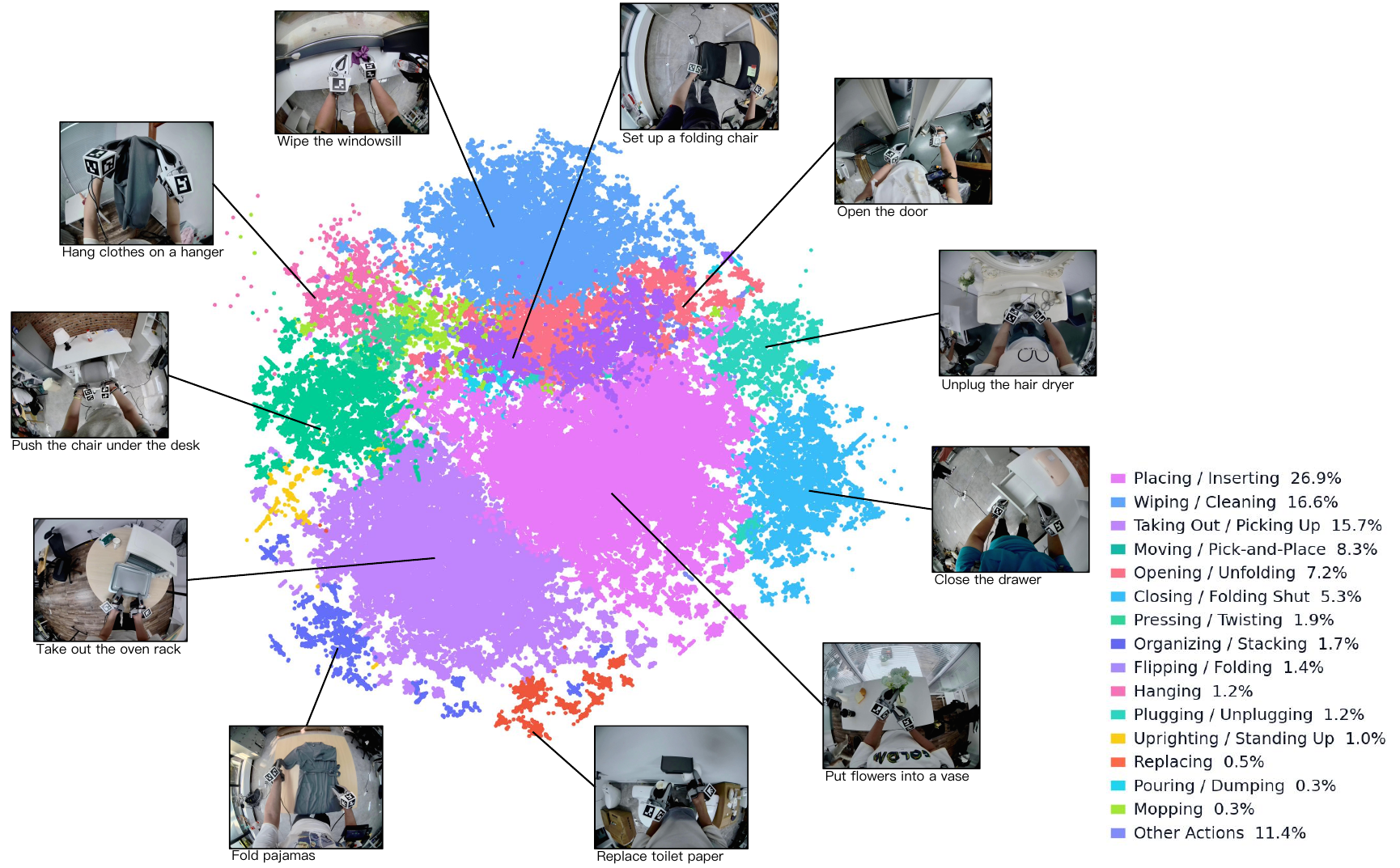}
  \caption{Overview of the dataset distribution across tasks, scenes, objects, and
  manipulation attributes.}
  \label{fig:data-distribution}
\end{figure}

Two properties beyond raw scale make this corpus usable for our study. First,
fidelity is enforced during collection and processing rather than audited
afterwards, so every retained episode clears the bar of \cref{tab:fidelity}.
Second, the composition of the corpus is measured rather than assumed:
\cref{fig:data-distribution} shows its distribution over tasks, scenes, objects,
and manipulation attributes, and the analysis-and-export stage of
\cref{sec:pipeline:processing} uses these statistics to control the proportions
of any exported training set. From the full corpus we curate and publicly
release \textbf{HiFi-UMI-2K}, a representative $2{,}000$-hour subset selected by
the same stage to balance task, scene, and attribute coverage. HiFi-UMI-2K
inherits the export format of the full corpus, so each released episode carries
the same synchronized six-view video, calibrated bimanual trajectories, gripper
states, language annotations, and subtask boundaries described above.
Versioning and per-sample quality metadata
(\cref{sec:pipeline:processing}) make every experimental subset reproducible and
traceable to its batch, device, and review history, linking performance to data
choices. HiFi-UMI-2K is distributed under the Creative Commons Attribution 4.0
International (CC BY 4.0)
license, which permits redistribution and derivative use, including for
commercial purposes, provided the source is attributed. Because the capture device records
head-mounted egocentric video, all human faces appearing in the released
recordings are masked before distribution, so HiFi-UMI-2K contains no facial
imagery. The dataset is intended for research on manipulation learning and on
robot data pipelines; we discourage its use for behavioral surveillance or
identity tracking.

\begin{table}[t!]
  \centering
  \caption{Dataset statistics for the full processed corpus (``Collected'') and
  curated HiFi-UMI-2K subset (``Released'').}
  \begin{tabularx}{\linewidth}{@{}Xr@{}}
    \toprule
    Property & Value \\
    \midrule
    \multicolumn{2}{@{}l}{\emph{Collected}} \\
    \quad Hours                   & $20{,}000+$ \\
    \quad Episodes                & $4{,}320{,}000+$ \\
    \quad Scenes                  & $480+$ \\
    \midrule
    \multicolumn{2}{@{}l}{\emph{Released} (HiFi-UMI-2K)} \\
    \quad Hours                   & $2{,}000$ \\
    \quad Episodes                & $482{,}100+$ \\
    \quad Scenes                  & $110+$ \\
    \midrule
    Camera views per episode      & $6$ \\
    \bottomrule
  \end{tabularx}
  \label{tab:dataset-stats}
\end{table}

\section{Baselines and Training Setup}
\label{sec:setup}

\paragraph{Design principle.}
Our goal is to test whether high-fidelity UMI data is sufficiently action-aligned
for post-training across reactive VLAs and WAMs, rather than for one policy
implementation. We instantiate three strong foundation-policy baselines:
\textbf{StarVLA-QwenPI}, \textbf{OpenPI-$\pi_{0.5}$}, and
\textbf{LingBot-VA}. We standardize the task definition, physical action
semantics, initial-state distribution, success criteria, and safety conditions;
each backbone retains its native observation, temporal sampling, action
tensorization, normalization, and deployment interface. Within each backbone,
we vary only the task-specific data source while holding the architecture,
initialization, optimization, and interfaces fixed. Agreement across backbones
is convergent evidence about the data, not a pooled architecture comparison.

\paragraph{Policy interfaces and physical action semantics.}
At a high level, each backbone consumes episodes of the form
\begin{equation}
    \tau =
    \left\{
    \left(
    o_t^{(m)}, q_t^{(m)}, \ell, a_{t:t+H_m-1}^{(m)}
    \right)
    \right\}_{t=1}^{T},
\end{equation}
where $m$ indexes the backbone, $o_t^{(m)}$ denotes its native visual
observation, $q_t^{(m)}$ denotes its native robot-state input, $\ell$ is the
natural-language task instruction, and $a_{t:t+H_m-1}^{(m)}$ is its native
future action tensor. The visual observation $o_t^{(m)}$ is drawn from the four
wrist views of the capture rig---two per hand---for every backbone and every
condition; the head-mounted stereo pair is used only for trajectory reconstruction during
capture and is never provided as input to the policy. Within each backbone, the UMI and
teleoperation variants use identical prompts, camera selection, temporal
offsets, tensor layout, normalization convention, and evaluation protocol.

At the physical robot interface, all three backbones use the same active
bimanual end-effector semantics. For arm $j$, every future pose in a chunk
beginning at $t_0$ is expressed relative to the same current-observation pose:
\begin{equation}
    \Delta\mathbf{T}_{t_0,h}^{j,(m)}
    =
    \left(\mathbf{T}_{t_0}^{j}\right)^{-1}
    \mathbf{T}_{t_0+\delta_h^{(m)}}^{j},
    \label{eq:chunk_anchor_action}
\end{equation}
where $\delta_h^{(m)}$ is the native future offset at index $h$ for backbone
$m$.
Thus, chunk rows share one measured anchor and are not defined recursively
relative to the preceding action target. Translation is expressed in the
anchor end-effector frame, orientation uses Rotation6D~\citep{rot6d}, and the
gripper target remains absolute. Each arm therefore contributes
$3+6+1=10$ active physical channels, giving 20 channels for bimanual tasks.
This common physical convention does not impose a common tensorization:
StarVLA-QwenPI uses the 20 channels directly, OpenPI-$\pi_{0.5}$ pads them
within its native 32-dimensional action tensor, and LingBot-VA maps them into
its native 30-dimensional tensor. Each backbone retains its own horizon,
padding, masking, temporal offsets, and normalization statistics.

\subsection{StarVLA-QwenPI}
\label{sec:setup:qwenpi}

StarVLA-QwenPI is our Qwen-based VLA baseline and follows the modular
backbone--action-head design of StarVLA~\citep{starvla}. We use
Qwen3-VL-4B-Instruct~\citep{qwen3technicalreport}, with 36 transformer layers and
a hidden width of 2{,}560, together with a $\pi$-style conditional
flow-matching DiT action head~\citep{pi0,flowmatching,dit}. Qwen encodes the
multi-view observation and instruction, and each DiT block cross-attends to the
corresponding layer-wise Qwen features.

The official QwenPI path lacks action-side self-attention, limiting token mixing
within predicted chunks. We retain cross-attention to all 36 Qwen layers and add
an attention-only self-attention residual after every odd-numbered DiT block,
coupling the $H=\texttt{20}$ action steps without repeating feed-forward
computation.

For a ground-truth action chunk $a$ and Gaussian noise
$\epsilon\sim\mathcal{N}(0,I)$, we sample
$u\sim\operatorname{Beta}(1.5,1.0)$ and set $\tau=(s-u)/s$, where $s=0.999$.
The action head is trained to predict the vector field from noise to data:
\begin{equation}
\begin{aligned}
    a^\tau &= (1-\tau)\epsilon+\tau a, \\
    \mathcal{L}_{\mathrm{FM}}
    &= \mathbb{E}\!\left[
        \left\|
        v_\theta(a^\tau,\tau,z_t)-(a-\epsilon)
        \right\|_2^2
    \right].
\end{aligned}
\end{equation}

At inference, we integrate the learned vector field from Gaussian noise using
\texttt{8} explicit-Euler steps. The resulting chunk contains
$H=\texttt{20}$ actions~\citep{act,openvlaoft}, of which the robot executes the
first $H_{\mathrm{exec}}=\texttt{10}$ in a receding-horizon
manner~\citep{diffusionpolicy}. Images are resized to
\texttt{224}$\times$\texttt{224}, and action dimensions are normalized using
statistics from the training split.

\subsection{OpenPI-$\pi_{0.5}$}
\label{sec:setup:pi05}

Our second VLA baseline, the JAX implementation of
OpenPI-$\pi_{0.5}$~\citep{pi05}, combines a
PaliGemma~\citep{paligemma} visual-language stream with a Gemma continuous action
expert. It uses a SigLIP So400m/14 visual encoder followed by an 18-layer
Gemma-2B language model, plus an 18-layer Gemma-300M action expert. The streams
have modality-specific parameters and Mixture-of-Transformers joint
attention~\citep{pi0}.

Images and the tokenized task instruction form the visual-language prefix.
Proprioceptive state is discretized and serialized together with the
instruction rather than injected as a continuous action-side token. Action
tokens attend to the complete prefix and to one another, allowing the full
action chunk to be predicted jointly.

We use the continuous flow-matching path released in OpenPI. Given conditioning
context $c=(o,q,\ell)$, action chunk $a$, Gaussian noise
$\epsilon\sim\mathcal{N}(0,I)$, and flow time $t$, we construct
$x_t=(1-t)a+t\epsilon$ and train the action expert to predict the velocity
field:
\begin{equation}
    \mathcal{L}_{\mathrm{FM}}
    =
    \mathbb{E}
    \left[
        \left\|
            v_\theta
            \left(
                x_t,t
                \mid
                c
            \right)
            -
            \left(
                \epsilon-a
            \right)
        \right\|_F^2
    \right].
\end{equation}
This is our sole fine-tuning objective; unlike the full $\pi_{0.5}$ recipe, we
omit knowledge insulation, autoregressive subtask generation, and auxiliary text
loss. At inference, OpenPI generates a continuous action chunk with 10 Euler
steps for the VLA receding-horizon executor
(\cref{sec:setup:deployment:vla}).

\subsection{LingBot-VA}
\label{sec:setup:lingbotva}

Our third baseline is LingBot-VA~\citep{lingbotva}, a causal WAM
that predicts future visual states before recovering the actions that realize
them. Let $z_t=E_{\mathrm{VAE}}(o_t)$ denote the latent of the synchronized
multi-view observation and $h_t=(z_{\leq t},a_{<t})$ the video--action history,
with $\ell$ denoting the task instruction. LingBot-VA factorizes joint
prediction as
\begin{equation}
\begin{aligned}
&p_\theta(a_{t:t+H-1},z_{t+1:t+K}\mid h_t,\ell) \\
&\quad =
p_\theta(a_{t:t+H-1}\mid z_{t+1:t+K},h_t,\ell)\,
p_\theta(z_{t+1:t+K}\mid h_t,\ell).
\end{aligned}
\end{equation}
The second factor predicts future video latents, while the first acts as an
inverse-dynamics model that decodes a continuous action chunk from the predicted
visual transition. LingBot-VA uses a block-causal mask to order the interleaved video–action chunks, ensuring that sequence modeling conforms to the factorization described above. Future video
latents and continuous actions are trained jointly with continuous flow
matching,
$\mathcal{L}_{\mathrm{LingBot}}
=\mathcal{L}_{\mathrm{video}}+\mathcal{L}_{\mathrm{action}}$, using equal loss
weights.

LingBot-VA follows the chunk-anchored physical convention in
\cref{eq:chunk_anchor_action}. Its 20 active bimanual dimensions are inserted
into the native 30-dimensional action tensor through a fixed channel map, and
unused channels are masked from the action loss. Rotation6D is formed from the
first two rows of the relative rotation matrix. Each profile reuses its
condition-specific normalization statistics throughout inference.

At deployment, receding-horizon prediction uses a bounded rolling KV cache,
video/action guidance scales of $5/1$, $8/16$ denoising steps, an attention
window of $24$, and the WAM protocol in \cref{sec:setup:deployment:wam}.

\subsection{Training Variants}
\label{sec:setup:variants}

We evaluate the following training conditions:

\begin{enumerate}
\item \textbf{Real-robot teleoperation post-training}. The policy is
post-trained on task-specific demonstrations collected directly on the target
robot through teleoperation. This setting represents the conventional
deployment-oriented training recipe, where the post-training data is fully
embodied in the target robot's observation space, action space, dynamics, and
gripper embodiment. We use it as the real-robot reference for comparison with
UMI post-training.

\item \textbf{UMI post-training}. The policy is post-trained using only our
high-fidelity UMI demonstrations for the target tasks. No real-robot
teleoperation trajectories are included in this setting. This is the central
experimental condition of the paper: it directly tests whether sufficiently
accurate robot-free demonstrations can serve not merely as pre-training data,
but as the sole post-training source for producing deployable manipulation
policies on a real robot.

\item \textbf{UMI pre-training + UMI post-training}. The policy first undergoes
continued pre-training on the full-scale UMI corpus, and is then post-trained
on the task-specific UMI subset. Both stages use robot-free UMI data, with no
real-robot teleoperation episodes. This variant tests whether broad
robot-free manipulation pre-training provides reusable visual-motor priors~\citep{datascaling}
that further improve downstream UMI-only specialization. We instantiate this
pre-training stage on StarVLA-QwenPI only; OpenPI-$\pi_{0.5}$ and LingBot-VA are
post-trained from their publicly released base checkpoints throughout.

\end{enumerate}

Within a given comparison, the variants share the same policy backbone and its
native action tensorization, observation interface, normalization convention,
temporal sampling, and deployment controller. Therefore,
performance differences mainly reflect the source and stage of the training
data rather than changes in model architecture or robot execution protocol. In
particular, comparing real-robot teleoperation
post-training against UMI post-training measures whether high-fidelity UMI data
can replace the conventional real-robot anchor, while comparing UMI post-training
against UMI pre-training plus UMI post-training measures the value of scaling
robot-free data before task-specific adaptation.

\subsection{Data Processing and Episode Filtering}
\label{sec:setup:data}

All UMI trajectories are first transformed into the deployment robot's
end-effector action convention. The reconstructed hand poses are expressed in a
shared world frame and converted into robot end-effector frames. Each
backbone-specific loader then selects its native future offsets
$\delta_h^{(m)}$ and constructs the chunk-anchored relative targets in
\cref{eq:chunk_anchor_action}. The resulting 20 active physical channels are
packed, padded, and masked according to the backbone's native tensor layout.
The teleoperation pipeline uses the same backbone-specific physical target
convention and tensor layout as its UMI counterpart.

Before backbone-specific conversion, we apply source-trajectory validity
checks. Because the corpus is already high-fidelity
(\cref{sec:pipeline:processed}), these checks rarely trigger on the UMI data.
Episodes are removed
if they contain any of the following: SLAM tracking failure, missing camera
frames, timestamp discontinuity, severe hand-pose outliers, action spikes above a
physical threshold, incomplete task execution, or inconsistent gripper-state
reconstruction. Long episodes are sliced into coherent segments with idle
prefixes and suffixes removed. Filtering precedes the train/validation split to
keep near-duplicate windows from crossing splits.

For normalization, each condition uses the statistics bound to its deployment
profile under the corresponding backbone's unchanged normalization convention.
Position, rotation, and gripper channels are normalized separately. For
bimanual tasks, left and right arms use separate statistics.

\subsection{Training and Optimization}
\label{sec:setup:optimization}

\paragraph{VLA pre-training.}
Pre- and post-training minimize the flow-matching behavior-cloning objective.
We optimize StarVLA-QwenPI end to end with
AdamW~\citep{adamw} ($\beta_1=0.9$, $\beta_2=0.95$,
$\epsilon=10^{-8}$, and weight decay $10^{-8}$) in mixed precision, with
gradient clipping at $1.0$ and no accumulation.

We pre-train on 4{,}000 hours of multi-task UMI data. One pass through this
training mixture corresponds to 180{,}000 optimization steps. For the scaling
and OOD analyses, the final exported checkpoint is obtained after a further
5{,}000-step linear learning-rate decay. Training uses an effective global batch size of 2{,}048. The schedule begins with a
3{,}000-step linear warm-up. Peak
learning rates are $10^{-4}$ for the action DiT and projection layers,
$2.5\!\times\!10^{-5}$ for the remaining backbone parameters, and $10^{-5}$
for the Qwen--action interface. We use one independently sampled flow time and
noise realization per training example, randomize the ordering of camera views,
and save checkpoints every 5{,}000 steps.

\paragraph{VLA post-training.}
For task-specific post-training of VLA models, the UMI-pretrained variant initializes all
parameters from
the 185k-step pre-training checkpoint. We train for 50{,}000 steps
with an effective global batch size of 512. After a
2{,}500-step linear warm-up, the learning rate follows cosine decay to
$7\!\times\!10^{-7}$; the peak rates are $5\!\times\!10^{-5}$ for the action
DiT and $10^{-5}$ for the Qwen backbone and Qwen--action interface. For each
task sample, we draw eight independent flow-time/noise realizations and average
their flow-matching losses. The Qwen-VL-initialized StarVLA-QwenPI baseline
instead begins task-specific training with a randomly initialized action policy,
as described in the post-training comparison below. For OpenPI-$\pi_{0.5}$, we
follow the OpenPI fine-tuning
pipeline and initialize from \texttt{pi05\_base} using the same converted UMI
data.

\paragraph{WAM post-training.}
For LingBot-VA, we follow the released task-specific post-training
implementation and initialize the video--action Transformer from
\texttt{lingbot-va-base}~\citep{lingbotva}. We update all Transformer
parameters while keeping the causal VAE and text encoder frozen. For each task
and data-source condition, we train for 3{,}500 steps with a global batch size
of 32. We use
AdamW~\citep{adamw} with $(\beta_1,\beta_2)=(0.9,0.95)$,
$\epsilon=10^{-8}$, and weight decay $0.01$, excluding bias, normalization, and
other one-dimensional parameters. The learning rate is linearly warmed up to $10^{-5}$ over
the first 25 steps, kept constant through step 3{,}000, and cosine-annealed to zero over
the final 500 steps. The video and action flow-matching losses are weighted
equally.

\paragraph{Validation and checkpoint selection.}
During validation, we report the held-out flow-matching loss, per-dimension
action error after de-normalization, gripper error, and rollout-level
metrics on the real robot. We do not select checkpoints solely by validation
loss. Instead, we choose checkpoints using a small fixed validation protocol and
then freeze them before the final blind real-robot evaluation.

\subsection{Deployment Protocol}
\label{sec:setup:deployment}

Across the two evaluation tracks, we standardize the task definition,
initial-state distribution, physical robot action semantics, safety wrapper,
workspace and velocity limits, and success and termination criteria. We
otherwise preserve each backbone's native temporal execution interface.
Consequently, UMI- and teleoperation-post-trained policies are compared under
strictly matched deployment settings within each backbone, without forcing the
VLA and WAM backbones into a common control cadence or chunk-consumption rule.

\subsubsection{VLA Deployment}
\label{sec:setup:deployment:vla}

Following the latency-matching principle of UMI~\citep{umi2024}, we deploy the
two VLA backbones with timestamped receding-horizon control. Sensor streams are
aligned to a common observation time,
\begin{equation}
t^{\mathrm{obs}}
=
t^{\mathrm{now}}
-
\max_s \tau_s^{\mathrm{obs}},
\end{equation}
where $\tau_s^{\mathrm{obs}}$ is the calibrated latency of sensor stream $s$.
Images, proprioception, and gripper state therefore correspond to the same
physical instant. For a query anchored at $t_0$, every row of the returned pose
chunk is restored independently from the synchronized query-time end-effector
pose according to \cref{eq:chunk_anchor_action}, rather than being recursively
integrated from the preceding target.

Predicted targets are assigned timestamps using the backbone-specific action
interval and submitted to the latency-compensated robot command buffer.
StarVLA-QwenPI uses the $H=20$, $H_{\mathrm{exec}}=10$ configuration described
above; OpenPI-$\pi_{0.5}$ retains its native fixed horizon and replanning
interval. For each backbone, all deployment parameters are held unchanged
between its UMI- and teleoperation-post-trained variants.

\subsubsection{WAM Deployment}
\label{sec:setup:deployment:wam}

LingBot-VA instead follows its native block-causal streaming protocol. After an episode reset, the first model call generates 12 executable actions, while subsequent calls predict a native two-block chunk containing 24 actions. For each matched UMI-versus-teleoperation comparison, we use the same execution schedule---12 actions after reset and 24 actions thereafter---for both post-training variants. At
the start of every executed chunk, we measure the current end-effector
poses and use them as the shared chunk anchors. Each predicted pose target is independently restored according to \cref{eq:chunk_anchor_action}, without accumulating action rows over time. The evaluated profiles retain a source-time stride of three, i.e., one source observation per three native action slots. Together with the VAE temporal downsampling factor of four, this gives 12 native action slots per latent video frame. The low-level robot controller tracks the resulting timestamped targets.

During execution, the client records one synchronized multi-view key observation every three executed actions. At each subsequent request, the action history from the preceding block and the newly collected real observations are first used to update LingBot-VA's bounded rolling KV cache. The next video--action chunk is then predicted from this updated context, keeping the causal history grounded in the robot's executed trajectory. The evaluated baseline performs synchronous inference at chunk boundaries, and the cache is reset whenever the episode or task prompt changes.

\section{Experiments}
\label{sec:experiments}

We design our experiments to answer a central question: can high-fidelity UMI
data serve as a post-training source for directly deployable real-robot
manipulation policies, without relying on real-robot teleoperation data? 
To separate the effect of data source from that of model
architecture, we conduct the comparison across three policy backbones spanning
both foundation-policy families: the VLA policies StarVLA-QwenPI and
OpenPI-$\pi_{0.5}$, and the WAM LingBot-VA.

\subsection{Experimental Design}
\label{sec:experiments:design}

\subsubsection{Evaluation Benchmark}
\label{sec:experiments:benchmark}
All real-robot experiments use a stationary bimanual platform with two
seven-joint Tianji Robotics Marvin M6 force-controlled arms. At deployment, the
platform matches the HiFi-UMI end effector: it carries the same gripper and four
wrist cameras described in \cref{sec:pipeline:hardware}. The head stereo pair is
used only for offline capture reconstruction and excluded from deployment, so
the policy receives a strict subset of the recorded views. Capture and
deployment thus have physically identical contact and observation interfaces;
the residual embodiment gap is confined to arm kinematics. Policies emit
end-effector pose targets, interpolated to $125$\,Hz and streamed to a controller
that solves inverse kinematics at $125$\,Hz and sends joint-angle commands over
EtherCAT at $1$\,kHz.

To ensure a fair comparison across policies, we follow established real-robot
evaluation protocols~\citep{barreiros2026careful} and conduct all rollouts through a
standardized benchmark system. Each rollout is jointly conducted by two
evaluators with separated responsibilities: a \emph{policy operator} loads and
launches the assigned policy, while a \emph{scene operator} independently
constructs the sampled test instance, including object placement and environment
configuration, according to the benchmark specification. This separation limits
operator bias and keeps initial conditions randomized and reproducible across
policies.

Before evaluation, we freeze the task definitions, object sets, language
instructions, policy checkpoints, initial-condition bank, task-specific
timeouts, and safety rules. For every rollout, task-relevant objects are newly
randomized within a predefined collision-free and robot-reachable tabletop
region. All policies are evaluated under the same initial-condition
distribution, while the policy order is randomized to reduce temporal and
operator bias. We conduct 40 rollouts for each task--policy pair.

The primary metric is binary task success. A rollout is successful only if the
complete task objective is achieved before the fixed timeout, the final state
remains stable for at least two seconds, and no safety intervention occurs.
A rollout terminates upon success, timeout, prolonged lack of task progress,
repeated ineffective motion, unrecovered object dropping, incorrect-object
interaction, or a safety stop. All termination reasons are recorded for failure
analysis. Hardware or communication faults not caused by the policy are marked
as invalid trials and repeated.

\subsubsection{Task Suite}
\label{sec:experiments:tasks}

\begin{figure*}[t]
  \centering

  \begin{subfigure}[t]{0.49\textwidth}
    \centering
    \includegraphics[width=\linewidth]{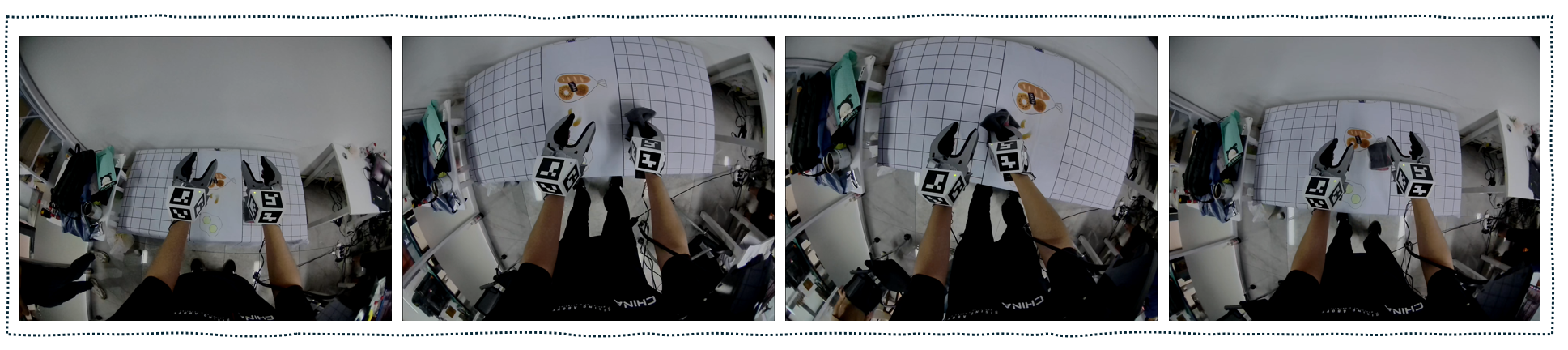}
    \caption{Stain Wiping}
    \label{fig:task_stain_wiping}
  \end{subfigure}
  \hfill
  \begin{subfigure}[t]{0.49\textwidth}
    \centering
    \includegraphics[width=\linewidth]{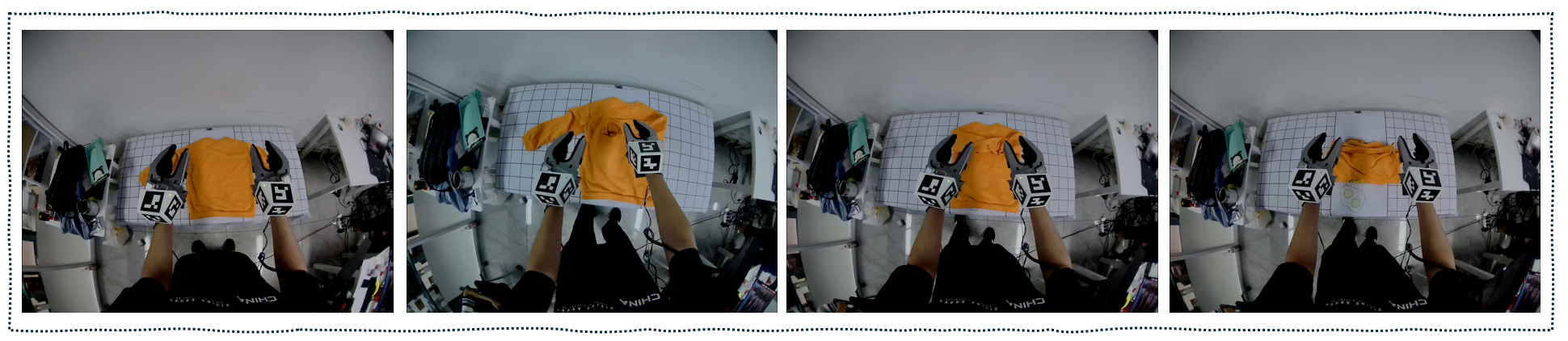}
    \caption{Shirt Folding}
    \label{fig:task_shirt_folding}
  \end{subfigure}

  \vspace{0.8em}

  \begin{subfigure}[t]{0.49\textwidth}
    \centering
    \includegraphics[width=\linewidth]{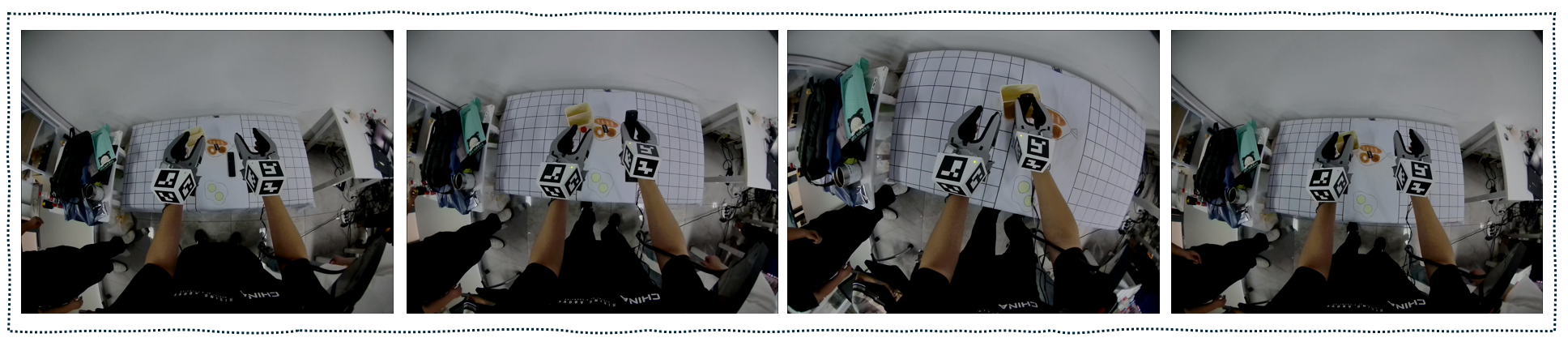}
    \caption{Remote Insertion}
    \label{fig:task_remote_insertion}
  \end{subfigure}
  \hfill
  \begin{subfigure}[t]{0.49\textwidth}
    \centering
    \includegraphics[width=\linewidth]{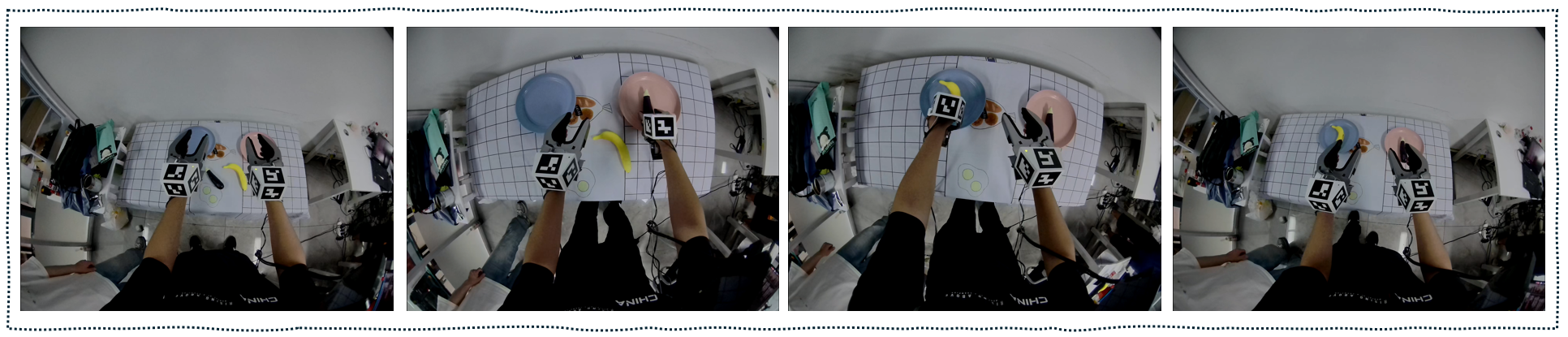}
    \caption{Produce Sorting}
    \label{fig:task_produce_sorting}
  \end{subfigure}

  \caption{The four benchmark tasks, shown as first-person HiFi-UMI
  demonstrations. Each panel shows representative stages of one task, captured
  from the head-mounted camera during handheld data collection. The suite spans
  contact-rich wiping, deformable-object manipulation, constrained placement, and
  semantic object sorting. Real-robot executions of the same tasks are shown in
  \cref{fig:rq1_real_robot_sequences}.
  }
  \label{fig:benchmark_tasks}
\end{figure*}

Our benchmark comprises four tabletop manipulation tasks that span contact-rich
interaction, deformable-object manipulation, constrained placement, and semantic
sorting (\cref{fig:benchmark_tasks}):

\begin{enumerate}
    \item \textbf{Stain Wiping}. The robot picks up a towel from the table, wipes
    the designated stain until the target region is clean, and finally places the
    towel back on the table.

    \item \textbf{Shirt Folding}. The robot folds the left and right sleeves of a
    shirt and then symmetrically folds the lower hem to produce the desired final
    configuration.

    \item \textbf{Remote Insertion}. The robot grasps a remote and inserts it
    into the target storage box.

    \item \textbf{Produce Sorting}. The robot places a designated vegetable onto
    the pink plate and a designated fruit onto the blue plate according to their
    semantic categories.
\end{enumerate}

The tasks span complementary capabilities: Stain Wiping tests sustained contact
and spatial coverage; Shirt Folding, bimanual deformable-object coordination;
Remote Insertion, precise constrained placement; and Produce Sorting, semantic
category-conditioned manipulation.

\subsubsection{Data Collection Regimes}
\label{sec:experiments:data}

We construct separate UMI and real-robot teleoperation datasets for every task.

\paragraph{UMI demonstrations.}
The UMI demonstrations are collected by multiple operators across different
physical sites, backgrounds, lighting conditions, and tabletop appearances. For
each task, we use 3{,}200 trajectories for post-training, corresponding to
approximately 10--20 hours of demonstrations depending on the task duration.
Because UMI collection does not require the target robot, independent operators
can collect data concurrently across multiple environments.

\paragraph{Real-robot teleoperation demonstrations.}
The teleoperation demonstrations are collected directly on one target robot in
the same physical environment used for evaluation. For each task, we collect
approximately 300 trajectories, corresponding to approximately 3--7 hours of
demonstrations depending on the task duration. Each collection session requires
both a teleoperator and an additional assistant for scene reset, object
placement, safety supervision, and recovery.

In our collection pipeline, obtaining one usable teleoperation trajectory
requires several times more wall-clock time than obtaining one UMI trajectory.
In addition to human operation time, real-robot collection incurs overhead from
robot execution, environment reset, safety checks, and hardware recovery. The
reported sizes therefore reflect practical pipeline throughput, not a
trajectory-count-matched design.

No UMI trajectory is collected in the evaluation scene. Consequently, the
UMI-post-trained policies are evaluated under a scene-level distribution shift
in background, illumination, tabletop appearance, and overall visual context. In
contrast, the teleoperation demonstrations are collected using the same robot
setup and environment as the evaluation. This asymmetry provides a conservative
test of whether diverse, multi-site UMI data can transfer to an unseen deployment
scene.

\subsubsection{Policy Comparisons and Evaluation Protocol}
\label{sec:experiments:protocol}

Our study varies three factors: the policy backbone, the source of
task-specific post-training data, and---for StarVLA-QwenPI only---the
initialization from which post-training starts. The design is deliberately
\emph{unbalanced}: the data-source axis is evaluated on all three backbones,
whereas the initialization axis is evaluated on one. \cref{tab:conditions}
enumerates the resulting seven training conditions and indicates where each is
reported.

\begin{table}[t]
  \centering
  \small
  \setlength{\tabcolsep}{4pt}
  \renewcommand{\arraystretch}{1.15}
  \caption{Seven training conditions. Each is evaluated on four tasks
  with $40$ rollouts per task. C1--C6 compare post-training sources within each
  backbone; C1 versus C7 compares initialization under matched HiFi-UMI
  post-training. C3--C6 use public base checkpoints, while only C7 adds HiFi-UMI
  pre-training (\cref{sec:setup:optimization}).}
  \label{tab:conditions}
  \begin{tabularx}{\linewidth}{@{}llXll@{}}
    \toprule
    & Backbone & Initialization / pre-training & Post-train data & Reported in \\
    \midrule
    C1 & StarVLA-QwenPI      & Qwen3-VL, scratch action head & HiFi-UMI      & \cref{sec:experiments:posttraining,sec:experiments:pretraining} \\
    C2 & StarVLA-QwenPI      & Qwen3-VL, scratch action head & Teleoperation & \cref{sec:experiments:posttraining} \\
    C3 & OpenPI-$\pi_{0.5}$  & \texttt{pi05\_base}           & HiFi-UMI      & \cref{sec:experiments:posttraining,sec:experiments:pretraining} \\
    C4 & OpenPI-$\pi_{0.5}$  & \texttt{pi05\_base}           & Teleoperation & \cref{sec:experiments:posttraining} \\
    C5 & LingBot-VA          & \texttt{lingbot-va-base}      & HiFi-UMI      & \cref{sec:experiments:posttraining} \\
    C6 & LingBot-VA          & \texttt{lingbot-va-base}      & Teleoperation & \cref{sec:experiments:posttraining} \\
    \midrule
    C7 & StarVLA-QwenPI      & Qwen3-VL $\rightarrow$ HiFi-UMI pre-training & HiFi-UMI & \cref{sec:experiments:pretraining} \\
    \bottomrule
  \end{tabularx}
\end{table}

Two comparisons follow from this design, each holding everything else fixed.
The \emph{data-source} comparison (C1~vs.~C2, C3~vs.~C4, C5~vs.~C6) fixes the
backbone, the initialization, the optimization recipe, the action
representation, the normalization protocol, and the deployment executor, and
changes only where the task-specific demonstrations come from. The
\emph{initialization} comparison (C1~vs.~C7) fixes the post-training data---the
same $3{,}200$ HiFi-UMI trajectories per task---together with the recipe and the
deployment stack, and changes only the checkpoint from which post-training
starts. Both arms of that comparison begin from the same Qwen3-VL weights, and
C7 differs only in that those weights are first carried through large-scale
HiFi-UMI pre-training. The comparison therefore isolates the visual-motor prior
acquired from HiFi-UMI data rather than the vision-language prior itself.
OpenPI-$\pi_{0.5}$ and LingBot-VA are post-trained from their publicly released
base checkpoints throughout and do not participate in the initialization
comparison.

For every condition--task pair, we conduct 40 real-robot rollouts. Before each
rollout, the manipulated objects are randomly repositioned within the permitted
workspace. Evaluation is performed on two identically configured robot platforms
under the same physical safety and task-evaluation protocol, while retaining
the backbone-specific execution clients described above. The primary metric is
the task success rate, $N_{\mathrm{success}}/N_{\mathrm{rollout}}$, under the
success and termination criteria fixed in \cref{sec:experiments:benchmark};
partial completion of the task objectives listed in
\cref{sec:experiments:tasks} is counted as failure.

\subsection{Can UMI-Only Post-Training Match Teleoperation-Based Post-Training?}
\label{sec:experiments:posttraining}

Across both evaluation tracks below, using HiFi-UMI rather than teleoperation as
the task-specific post-training source yields approximate aggregate parity under
the evaluated setting. Taking UMI minus teleoperation, the aggregate success-rate
differences are $-2.5$, $+3.1$, and $-0.6$ percentage points for
StarVLA-QwenPI, OpenPI-$\pi_{0.5}$, and LingBot-VA, respectively, with no
consistent direction across backbones. Because the VLA and WAM tracks use
different model-specific experimental settings, including temporal, training,
and deployment protocols, their absolute success rates are not directly
comparable. We therefore interpret their agreement as convergent evidence from
three controlled within-backbone comparisons rather than as a pooled comparison
across policy paradigms. We report the two VLA backbones and the WAM in turn.

\begin{figure*}[t]
  \centering

  \begin{subfigure}[t]{0.49\textwidth}
    \centering
    \includegraphics[width=\linewidth]{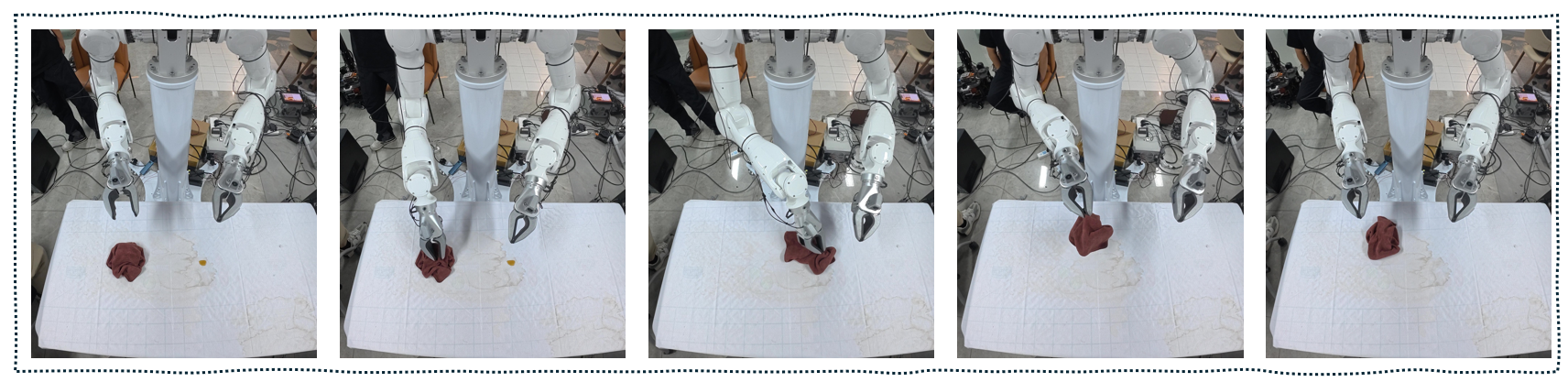}
    \caption{Stain Wiping}
    \label{fig:task_stain_wiping_robot}
  \end{subfigure}  
  \hfill
  \begin{subfigure}[t]{0.49\textwidth}
    \centering
    \includegraphics[width=\linewidth]{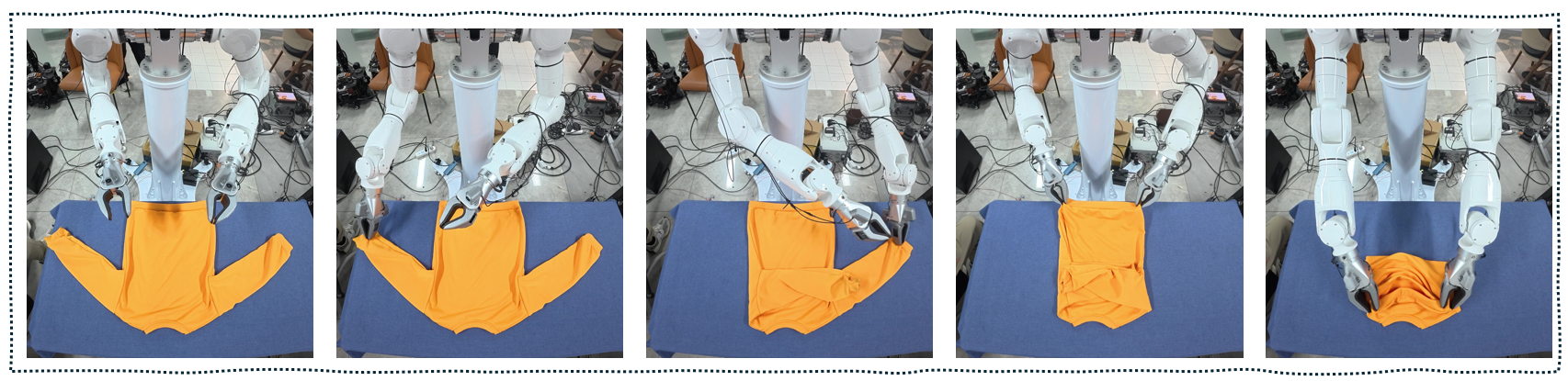}
    \caption{Shirt Folding}
    \label{fig:task_shirt_folding_robot}
  \end{subfigure}

  \vspace{0.8em}

  \begin{subfigure}[t]{0.49\textwidth}
    \centering
    \includegraphics[width=\linewidth]{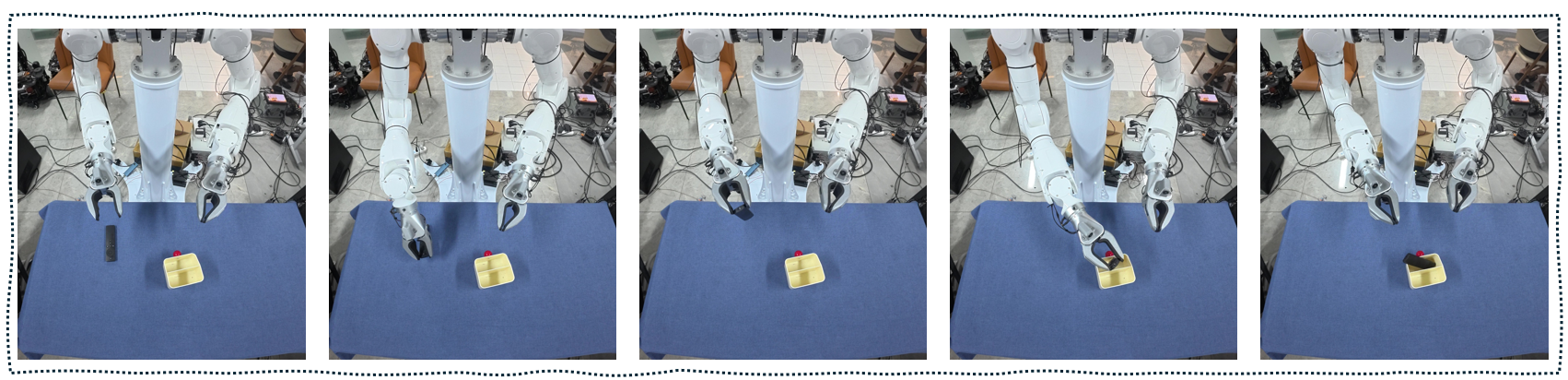}
    \caption{Remote Insertion}
    \label{fig:task_remote_insertion_robot}
  \end{subfigure}
  \hfill
  \hfill
  \begin{subfigure}[t]{0.49\textwidth}
    \centering
    \includegraphics[width=\linewidth]{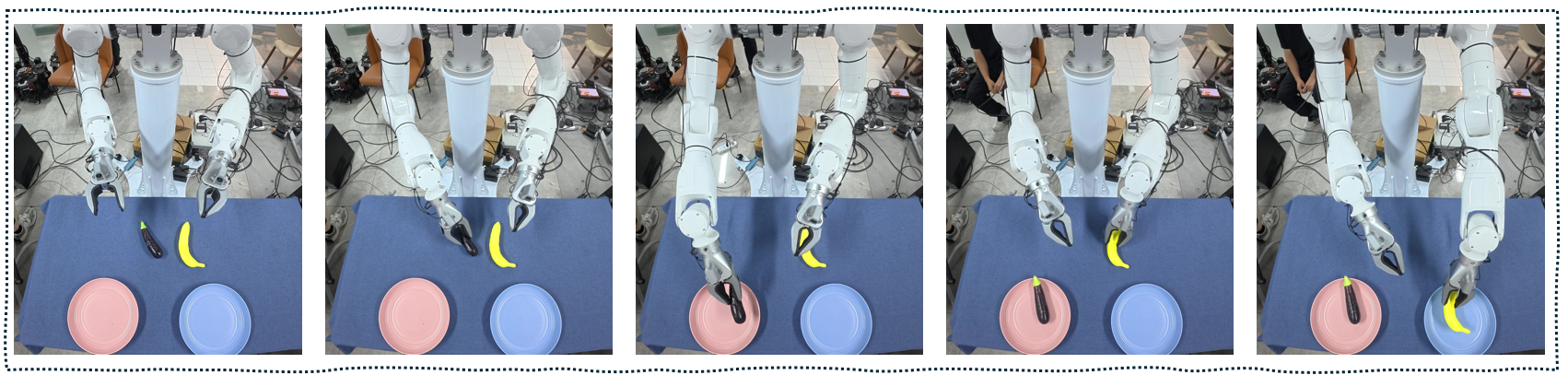}
    \caption{Produce Sorting}
    \label{fig:task_produce_sorting_robot}
  \end{subfigure}

  \caption{
    Representative successful real-robot executions of the four benchmark tasks, with time progressing from left to right in each row. 
  }
  \label{fig:rq1_real_robot_sequences}
\end{figure*}

\subsubsection{Evaluation Track I: VLA Backbones}

We compare UMI-only post-training against conventional real-robot
teleoperation post-training using two VLA backbones. Within each backbone,
the model architecture, initialization, optimization recipe, action
representation, and deployment stack are held fixed; the task-specific 
post-training dataset is supplied by either the HiFi-UMI or teleoperation 
collection pipeline, while the model and deployment stack are held fixed 
within each backbone. In total, this evaluation comprises
640 real-robot rollouts across four tasks, four policy variants, and 40 trials
per task--policy pair. Figure~\ref{fig:rq1_real_robot_sequences} shows successful real-robot rollouts
for all four tasks.

\begin{figure*}[t]
  \centering
  \includegraphics[width=\textwidth]
  {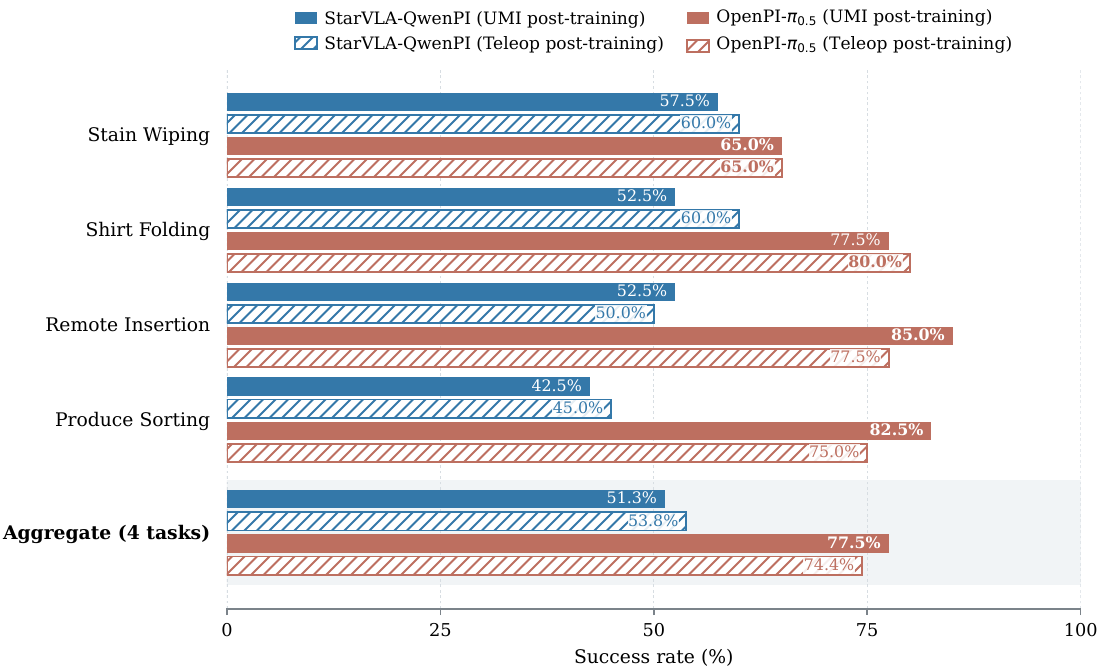}
    \caption{Real-robot success rates across the two VLA backbones and
    post-training data sources. Each task-policy pair is evaluated over 40 rollouts, while the
    aggregate result pools all four tasks (160 rollouts per policy). Solid bars
    denote UMI post-training and hatched bars denote real-robot teleoperation
    post-training. Bold labels indicate the highest success rate within each task.}
  \label{fig:posttraining_success}
\end{figure*}

\paragraph{Aggregate comparison.}
As shown in Fig.~\ref{fig:posttraining_success}, UMI-only post-training closely matches
teleoperation-based post-training for both VLA backbones. On
StarVLA-QwenPI, the UMI-post-trained policy achieves 51.3\% success
(82/160), compared with 53.8\% (86/160) for its teleoperation-trained
counterpart, corresponding to a difference of only 2.5 percentage points.
On OpenPI-$\pi_{0.5}$, UMI post-training achieves 77.5\% success
(124/160), exceeding teleoperation post-training at 74.4\% (119/160) by
3.1 percentage points. Pooled, UMI yields 206/320 (64.4\%) versus 205/320
(64.1\%) for teleoperation; within-backbone comparisons remain primary.

Notably, the aggregate UMI--Teleop gap remains small for both VLA backbones,
whereas the difference between the two backbones is substantially
larger. This indicates that the conclusion does not depend on a particular VLA
architecture: replacing teleoperation demonstrations with high-fidelity UMI
demonstrations does not systematically reduce real-robot performance.

\paragraph{Task-level comparison.}
The direction of the UMI--Teleop difference varies across tasks rather than
consistently favoring either data source. On Stain Wiping, the two conditions
perform similarly for StarVLA-QwenPI and achieve an exact tie at 65.0\% for
OpenPI-$\pi_{0.5}$. Teleoperation provides a modest advantage on Shirt
Folding, reaching 60.0\% versus 52.5\% with StarVLA-QwenPI and 80.0\% versus
77.5\% with OpenPI-$\pi_{0.5}$.
Conversely, UMI post-training is slightly higher on Remote Insertion:
StarVLA-QwenPI reaches 52.5\% versus 50.0\% with teleoperation, and
OpenPI-$\pi_{0.5}$ reaches 85.0\% versus 77.5\%. On Produce Sorting, OpenPI also
yields a higher rate with UMI (82.5\% versus 75.0\%). These differences may
reflect UMI's broader variation in objects, backgrounds, and collection sites,
which can aid spatial grounding and object-conditioned control.

Because each task--policy pair contains 40 rollouts, one successful trial
corresponds to 2.5 percentage points. Most task-level differences therefore
represent only one to three rollouts. We consequently interpret small differences as
approximate parity rather than decisive superiority of one data source.

\paragraph{Generalization under scene shift.}
This comparison is conservative for UMI with respect to evaluation-scene visual 
shift, but it is not sample matched. The UMI demonstrations
are collected by multiple operators across environments that differ from the
evaluation scene in background, illumination, tabletop appearance, and spatial
layout. In contrast, the teleoperation demonstrations are collected directly on
the target robot in the same environment used for evaluation. Despite this
scene-level distribution shift, UMI-post-trained policies maintain aggregate
performance comparable to their in-environment teleoperation counterparts. This
is consistent with diverse, high-fidelity UMI data providing accurate
supervision under scene shift.

Across the two VLA backbones, the results provide evidence that high-fidelity
UMI data can serve as the sole task-specific post-training source for directly
deployable VLA policies, without requiring a real-robot teleoperation anchor.
The comparison is not sample matched: each task is post-trained on 3{,}200
UMI trajectories but 300 teleoperation trajectories. We therefore interpret
these experiments as a comparison between practical data-production pipelines,
rather than as a claim of equal-sample data efficiency. Under this practical
collection regime, however, removing real-robot teleoperation from
post-training does not produce a corresponding loss in deployment performance.

\paragraph{How Much UMI Data Is Needed for VLA Deployment?}

While the preceding VLA comparisons demonstrate that UMI-only post-training can
match teleoperation-based post-training, an important practical question remains:
\emph{how much UMI data is required to obtain a deployable manipulation policy?}
To answer it, we study how UMI demonstration count affects real-robot success.

We select the Remote Insertion task for this analysis. This task provides a
controlled yet challenging evaluation setting: it requires a sequence of
manipulation skills, including object grasping, transportation, precise pose
alignment, and insertion into a constrained target region. Compared with highly
deformable-object tasks such as shirt folding, Remote Insertion has lower
intrinsic variance, allowing us to better isolate the effect of demonstration
quantity on policy learning.

Following recent studies on data scaling in robotic imitation learning
~\citep{datascaling}, we train the same OpenPI-$\pi_{0.5}$ backbone using
different subsets of our UMI demonstrations. Specifically, we construct five
training sets containing 400, 800, 1{,}600, 3{,}200, and 6{,}400 episodes,
respectively. All other training configurations, including model initialization,
optimization settings, action representation, and evaluation protocol, are kept
identical. Each trained policy is evaluated on the real robot with 40
independent rollouts under randomized initial object configurations.

\begin{figure}[t]
  \centering
  \includegraphics[width=0.98\linewidth]
    {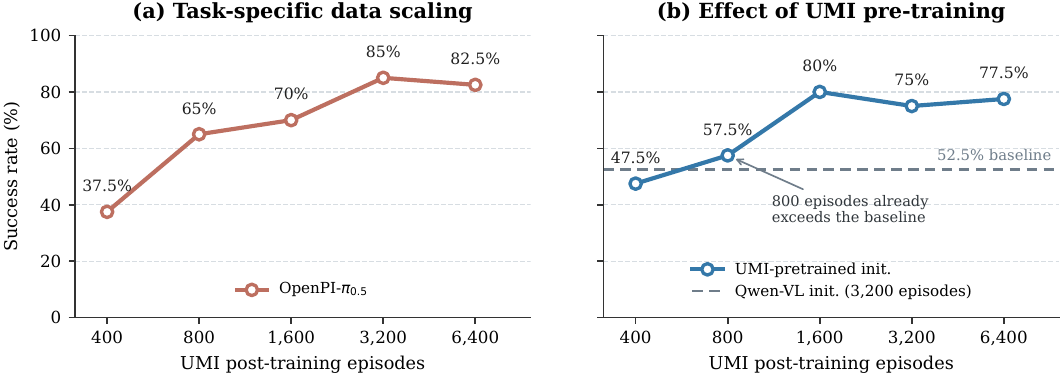}
  \caption{Two complementary Remote Insertion scaling studies.
  (a)~Task-specific data scaling for OpenPI-$\pi_{0.5}$ post-trained on
  increasingly large UMI subsets. (b)~Initialization ablation for
  StarVLA-QwenPI post-training; the dashed line marks the Qwen-VL-initialized
  scratch action policy trained on 3{,}200 episodes. The UMI-pretrained policy
  exceeds this baseline with only 800 episodes. Each setting is evaluated over
  40 real-robot rollouts.}
  \label{fig:umi_scaling}
\end{figure}

As shown in Fig.~\ref{fig:umi_scaling}a, increasing the amount of UMI
post-training data leads to substantial performance improvements in the
low-data regime. The success rate improves from 37.5\% with 400 demonstrations
to 65.0\% with 800 demonstrations, indicating that additional UMI trajectories
rapidly improve the policy's ability to acquire the basic manipulation skill.
Further scaling to 1{,}600 and 3{,}200 demonstrations continues to improve
performance, reaching 70.0\% and 85.0\% success rates, respectively.

However, after approximately 3{,}200 demonstrations, the performance improvement
largely saturates. Increasing the training set size from 3{,}200 to 6{,}400
episodes does not provide additional gains, with the success rate slightly
decreasing from 85.0\% to 82.5\%. This suggests that performance plateaus by
3{,}200 episodes at the resolution of our 40-rollout evaluation.

Together with the previous UMI-versus-teleoperation comparison, these results
support the conclusion that high-fidelity UMI data provides scalable
task-specific supervision. Rather than requiring a small amount of expensive
robot-collected ``anchor'' data, increasing robot-free demonstrations substantially 
improves task-specific post-training in the low-data regime and reaches competitive 
deployment performance without a real-robot teleoperation anchor.

\subsubsection{Evaluation Track II: WAM Policy}

We conduct a separate UMI-versus-teleoperation comparison within
LingBot-VA, a WAM that generates actions through predicted future
video and inverse-dynamics decoding. We construct UMI- and
teleoperation-post-trained variants from the same LingBot-VA base checkpoint,
while keeping the model architecture, optimization schedule, video--action
objective, action representation, inference settings, and deployment stack
fixed. The task-specific post-training data are supplied by the corresponding
UMI or teleoperation collection pipeline. For each task--policy pair, we conduct
40 real-robot rollouts, yielding 320 rollouts across four tasks and two policy
variants. Because this WAM evaluation follows a model-specific training and
evaluation protocol, its absolute success rates are not directly compared with
those of the VLA track. As an offline diagnostic, we additionally evaluate action prediction while
conditioning on ground-truth future video. This oracle conditioning removes
accumulated future-video generation error from the diagnostic and allows us to
probe the inverse-action component more directly.

\paragraph{Aggregate comparison.}

Across all four tasks, UMI-only post-training achieves an aggregate success rate of 56.9\% (91/160), closely matching the 57.5\% (92/160) obtained with real-robot teleoperation post-training. This comparable performance suggests that the two data sources provide similar effectiveness for post-training under the evaluated setting, rather than indicating a clear advantage of either source. Under the separate WAM evaluation protocol, replacing teleoperation demonstrations with HiFi-UMI does not result in a systematic degradation in closed-loop deployment performance.

\begin{figure*}[t]
  \centering
  \includegraphics[width=0.95\textwidth]
  {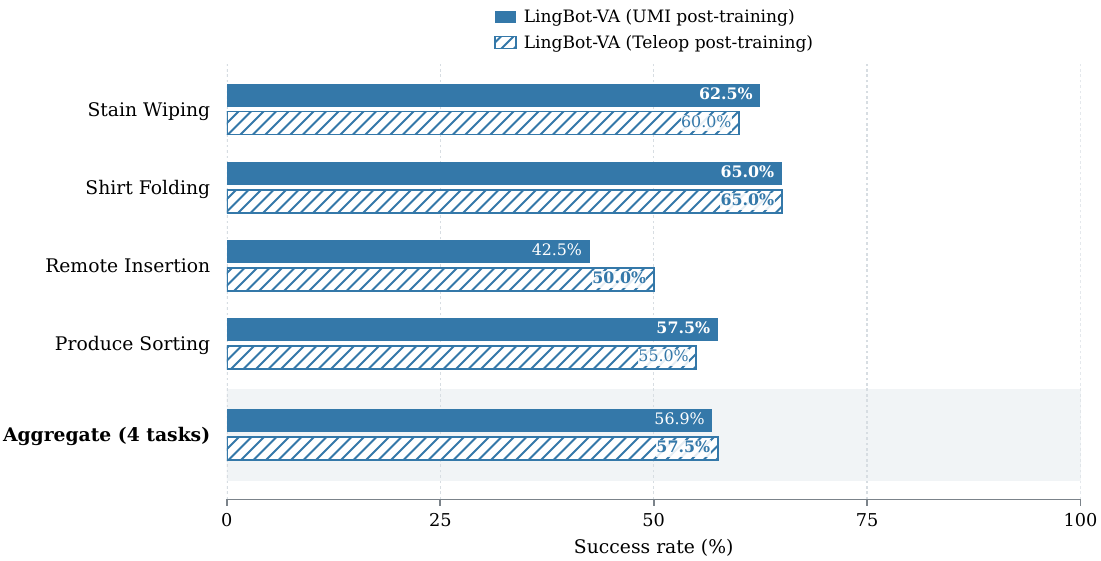}
    \caption{Real-robot success rates for LingBot-VA under UMI versus
    teleoperation post-training. Each task--data-source pair is evaluated over
    40 rollouts, while the aggregate result pools all four tasks (160 rollouts
    per post-training source). Solid bars denote UMI post-training and hatched
    bars denote real-robot teleoperation post-training. Bold labels indicate the
    highest success rate within each task.}
  \label{fig:wam_posttraining_success}
\end{figure*}

\paragraph{Task-level comparison.}
The direction of the success-rate difference varies across tasks rather than consistently favoring either data source. UMI post-training performs slightly better on the Stain Wiping task, achieving 62.5\% success compared with 60.0\% for teleoperation, and similarly reaches 57.5\% on the Produce Sorting task compared with 55.0\%. The two variants achieve the same success rate of 65.0\% on the Shirt Folding task. Conversely, teleoperation post-training performs better on the Remote Insertion task, reaching 50.0\% success compared with 42.5\% for UMI post-training. With 40 rollouts per task--policy pair, these small variations do not consistently favor either source.

Beyond binary success, qualitative inspection of the robot rollouts reveals a
systematic difference in execution tempo: UMI-post-trained policies tend to
produce larger-amplitude, more continuous, and more natural-looking motions.
Successful rollouts often reach task milestones with fewer pauses, whereas the
teleoperation-post-trained policies more often execute incremental corrections.
These observations are qualitative, as completion time is also influenced by recovery behaviors and termination conditions.

More direct nominal execution does not necessarily imply stronger recovery behavior. This distinction is most evident on the Remote Insertion task: the UMI-post-trained policy often performs a decisive initial grasp, but may require multiple attempts under imperfect contact conditions. These retries reduce the overall execution efficiency and are consistent with the lower success rate observed on this task. This result highlights the distinction between nominal execution efficiency and recovery robustness, suggesting that broader coverage of contact correction and regrasp behaviors remains important for reliable closed-loop deployment.

\FloatBarrier
\paragraph{Ground-truth-video analysis shows stable cross-domain pose decoding.}

The closed-loop evaluation above measures the performance of the full WAM pipeline, but it does not isolate whether failures arise from future-video prediction or action decoding. As LingBot-VA's action prediction is conditioned on
predicted future visual latents, evaluation with generated video combines
visual-generation and action-decoding errors. To disentangle these factors, we replace the
generated future-video latents with cached ground-truth latents and evaluate
the action decoder on the Stain Wiping, Shirt Folding, Remote Insertion, and
Produce Sorting tasks. This
\emph{ground-truth-video} protocol serves as an oracle diagnostic rather than a deployable
inference mode: it removes future-video generation error from the evaluation,
but does not by itself establish that video generation is the dominant
closed-loop bottleneck. We evaluate three settings: a teleoperation-post-trained model on
real-robot held-out observations (Real$\rightarrow$Real), a UMI-post-trained
model on the same held-out domain (UMI$\rightarrow$Real), and the same
UMI-post-trained model on episode-disjoint UMI held-out observations
(UMI$\rightarrow$UMI).

\begin{figure*}[t]
  \centering
  \includegraphics[width=0.98\textwidth]
  {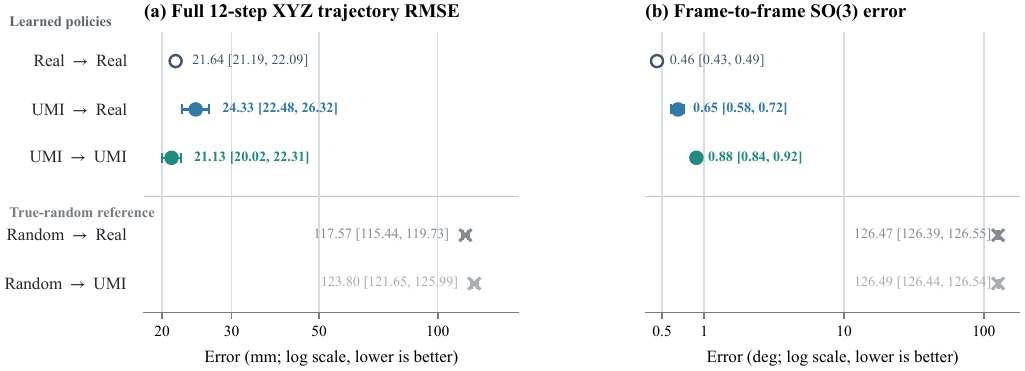}
  \caption{WAM pose-decoding fidelity under ground-truth-video
  conditioning. Points denote equal-task means across Stain Wiping, Shirt
  Folding, Remote Insertion, and Produce Sorting; whiskers denote 95\%
  source-episode bootstrap confidence intervals.
  Translation is the bimanual XYZ RMSE over the complete first $H=12$
  chunk in the native chunk-anchor representation. Rotation is the SO(3)
  geodesic error between predicted and ground-truth adjacent-frame increments
  over $t=2,\ldots,12$. Logarithmic axes preserve resolution among learned
  policies while retaining true-random actions as a common scale reference.
  Gripper channels are excluded because their physical semantics are not
  calibrated across the UMI and real-robot interfaces. Lower is better.}
  \label{fig:wam_gtvideo_pose_fidelity}
\end{figure*}

\paragraph{Metrics and aggregation.}
We evaluate the complete first executed action chunk, with horizon $H=12$.
After applying the same de-normalization and chunk-anchor SE(3) reconstruction
to predictions and targets, we compute the bimanual XYZ trajectory error as
\begin{equation}
  E_{\mathrm{XYZ}}
  =
  10^3
  \sqrt{
    \frac{1}{6H}
    \sum_{t=1}^{H}
    \sum_{b\in\{L,R\}}
    \left\|
      \hat{\mathbf{p}}_{t,b}-\mathbf{p}_{t,b}
    \right\|_2^2
  },
  \label{eq:wam-xyz-rmse}
\end{equation}
where the factor $10^3$ converts meters to millimeters. Evaluating the complete
chunk preserves the native chunk-anchor action contract and measures the whole
trajectory that is passed to the controller, rather than a selected shorter
prefix. For rotation, we first convert Rotation6D predictions and targets to valid
rotation matrices and form strict adjacent-frame increments,
$\Delta\mathbf{R}_{t,b}=\mathbf{R}_{t-1,b}^{\top}\mathbf{R}_{t,b}$. We then
report
\begin{equation}
  E_{\mathrm{rot}}
  =
  \frac{1}{2(H-1)}
  \sum_{t=2}^{H}
  \sum_{b\in\{L,R\}}
  d_{\mathrm{SO(3)}}\!\left(
    \widehat{\Delta\mathbf{R}}_{t,b},
    \Delta\mathbf{R}_{t,b}
  \right),
  \label{eq:wam-frame-rotation}
\end{equation}
in degrees, where $d_{\mathrm{SO(3)}}$ is the geodesic angle between two
rotations. Step $t=1$ is omitted because its preceding pose is the external
chunk anchor rather than a previous target in the predicted sequence. Adjacent
increments therefore measure local rotation direction and magnitude without
allowing accumulated anchor-relative drift to dominate the metric.

For the learned policies, we first average evaluation records and the three
inference seeds within each source episode. For the random references, we
instead average 256 independently sampled action chunks per record before
episode aggregation. We then average source episodes within each task and
assign equal weight to the four tasks. The reported 95\% confidence intervals
use 20{,}000 stratified bootstrap resamples with source episode as the sampling
unit, avoiding the treatment of correlated windows or stochastic samples as
independent observations. For each evaluation record, the random reference
samples every time step and each of the 20 active action channels independently
from $\mathrm{Uniform}[-1,1]$ in the corresponding task's frozen UMI
checkpoint normalization space, applies that checkpoint's inverse quantile
normalization, and passes the resulting chunk through the same physical
reconstruction and scoring pipeline. The primary analysis excludes the two gripper channels because their
acquisition semantics are not calibrated across interfaces. In particular,
when no grasp is present, real-robot teleoperation defaults to a fully open
gripper, whereas UMI retains the operator's hand-gesture angle. Including the
channels would therefore conflate pose-decoding fidelity with an interface
difference in gripper behavior.

As shown in Figure~\ref{fig:wam_gtvideo_pose_fidelity}, the UMI-post-trained policy achieves a full-chunk XYZ RMSE of 24.33 mm and a frame-to-frame SO(3) error of $0.65^\circ$ on held-out real-robot observations. On episode-disjoint held-out UMI observations, the corresponding errors are 21.13 mm and $0.88^\circ$, resulting in cross-domain differences of only 3.20 mm in translation and $0.23^\circ$ in rotation. On the same held-out real-robot domain, the teleoperation-post-trained reference achieves 21.64 mm XYZ RMSE and $0.46^\circ$ SO(3) error. For comparison, true-random actions yield 117.57\ mm and $126.47^\circ$ on
real-robot observations, and 123.80\ mm and $126.49^\circ$ on UMI
observations. Relative to the domain-matched Random$\rightarrow$Real
reference, UMI$\rightarrow$Real reduces translation and rotation error by
79.3\% and 99.5\%, respectively. These results indicate that, when future-video generation is bypassed, both UMI- and teleoperation-post-trained pose decoders achieve comparable centimeter-level translation accuracy and sub-degree local rotation accuracy. The small gap between UMI$\rightarrow$Real and UMI$\rightarrow$UMI further suggests that the UMI-trained decoder generalizes across observation domains without substantial degradation.
\FloatBarrier

\subsection{Does Large-Scale UMI Pre-Training Yield a Better Base Model?}
\label{sec:experiments:pretraining}

The two post-training tracks show that high-fidelity UMI demonstrations can
replace real-robot teleoperation as the task-specific data source. We next
isolate a separate question on StarVLA-QwenPI: whether the same robot-free data
can support a reusable base model. We pre-train StarVLA-QwenPI on 4{,}000 hours
of multi-task UMI data. A useful initialization should improve not only
prediction on held-out samples from this corpus, but also transfer to unseen
tasks and real-robot post-training. We study these properties in turn.

\paragraph{Scaling on held-out data.}
The 4{,}000-hour training mixture is drawn from the large-scale corpus described
in Sec.~\ref{sec:dataset} and spans diverse scenes, objects, and manipulation
skills. We reserve a fixed set of action chunks for evaluation and use the same
fixed-step Euler integration procedure for the flow-matching policy at every
checkpoint. This isolates the effect of greater training exposure from changes in
the evaluation data or sampler. One pass through the mixture ends at 180k steps,
followed by a short learning-rate decay.

\begin{figure}[t!]
  \centering
  \includegraphics[width=0.98\linewidth]{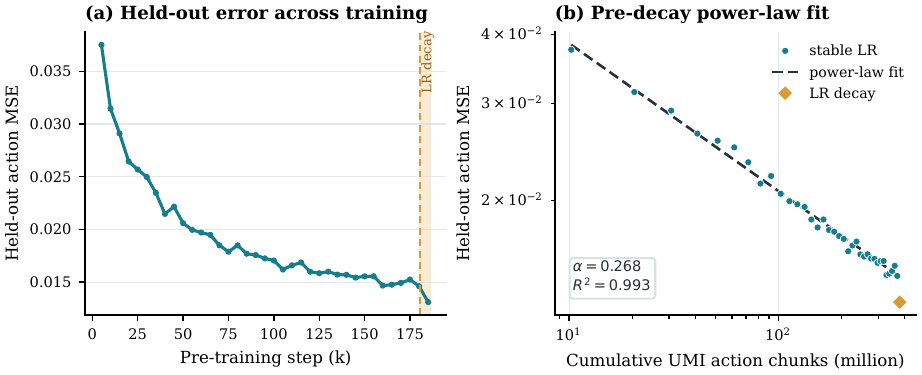}
  \caption{Held-out action-prediction error during large-scale UMI pre-training.
  (a)~Error across training, with the learning-rate decay shaded.
  (b)~Pre-decay checkpoints on logarithmic axes with a power-law fit. Because the
  corpus is fixed, the fit measures exposure scaling rather than dataset-size
  scaling.}
  \label{fig:pretraining_scaling}
\end{figure}

Figure~\ref{fig:pretraining_scaling} shows that held-out action error falls
by 61\% over one pass through the corpus. At a fixed model size and training
recipe, we fit
\begin{equation}
    \mathcal{L}_{\mathrm{heldout}}(S)
    = \mathcal{L}_{\infty} + A S^{-\alpha},
\end{equation}
where $S$ denotes the cumulative number of UMI action chunks processed globally.
The fitted exponent is $\alpha=0.268$, with $R^2=0.993$ before learning-rate
decay. This close log--log fit shows that the model continues to convert greater
UMI exposure into better action prediction throughout the training pass. The
trend is therefore not an artifact of the final decay schedule.

\paragraph{Transfer to unseen tasks.}
Lower error on the pre-training distribution does not by itself imply broader
visual-motor competence. We therefore construct a balanced evaluation set from
ten separately collected manipulation tasks that are absent from pre-training.
The same action chunks and inference settings are used for every checkpoint, and
the scaling fit uses only checkpoints before learning-rate decay.

\begin{figure}[!htbp]
  \centering
  \includegraphics[width=0.98\linewidth]{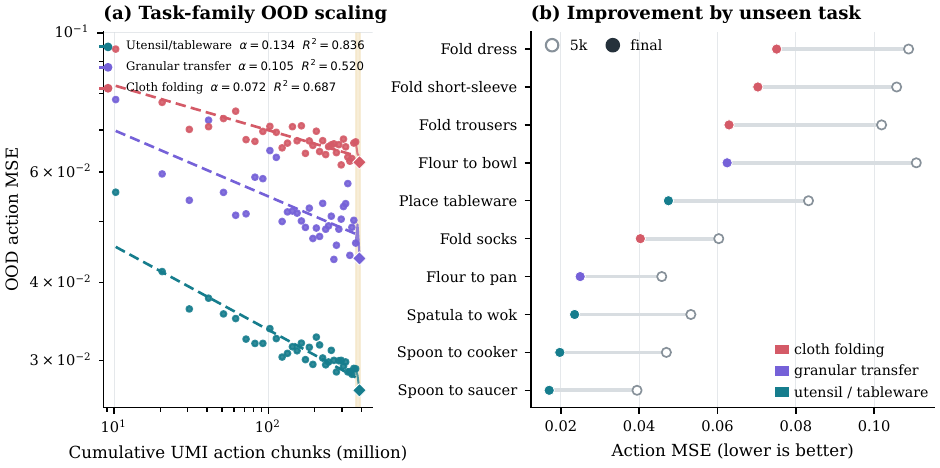}
  \caption{Generalization to ten unseen UMI tasks. (a)~Mean action MSE by
  manipulation family with pre-decay power-law fits; diamonds denote the final
  model export. (b)~Task-level error at the first checkpoint and final export.
  Colors identify broad manipulation families.}
  \label{fig:ood_generalization}
\end{figure}

Figure~\ref{fig:ood_generalization}a shows a 41\% reduction in mean OOD
error, and every unseen task improves. Aggregate OOD error also follows a
power-law trend, although with a smaller exponent of $\alpha=0.095$. The family
curves reveal a clear ordering: utensil and tableware interactions improve
fastest, granular transfer lies in the middle, and cloth folding improves more
slowly. Scaling is therefore shared across task families, but its rate depends on
which interaction patterns are represented during pre-training.

The task-level comparison in Fig.~\ref{fig:ood_generalization}b makes this
dependence more concrete. Rigid utensil-to-receptacle tasks have the lowest final
error, while garment folding remains the most difficult. Granular transfer also
changes markedly with target geometry. This structure mirrors the pre-training
mixture: object and receptacle placement accounts for more than one third of its
frames, while textile folding accounts for less than one percent. The model can
therefore reuse abundant rigid-object pick-and-place experience for novel
utensils, whereas cloth folding must extrapolate from sparse deformable-object
supervision. OOD transfer is governed more by coverage of interaction dynamics
than by whether the test object itself has appeared before.

This analysis also gives a concrete data-collection priority. Deformable-object
data should cover more garment topologies, initial configurations, bimanual
regrasps, and fold transitions. Granular-material data should vary container
geometry, fill level, and motion type. The OOD benchmark thus serves as a
diagnostic for rebalancing the next pre-training mixture, rather than only as a
single aggregate score.

\paragraph{Benefits for post-training.}
Offline action error is useful only if it predicts a better starting point for
deployment. On Remote Insertion, we compare with a Qwen-VL-initialized model
whose action policy starts from scratch; the shared visual-language backbone
isolates UMI's visual-motor prior.
With only 800 task-specific episodes, the UMI-pretrained policy already exceeds
the Qwen-VL-initialized baseline trained on four times as much data, as shown in
Fig.~\ref{fig:umi_scaling}b. Scaling to 1{,}600 episodes raises success to 80\%,
and the advantage remains at the matched 3{,}200-episode scale. The important
pattern is that strong performance appears earlier and remains above the
task-specific baseline: pre-training improves both data efficiency and the
performance reached after post-training.

We next test whether this advantage extends beyond a single task. On each of the
four benchmark tasks from Sec.~\ref{sec:experiments:tasks}, we post-train two
StarVLA-QwenPI policies using the same 3{,}200 task-specific trajectories. One
starts from Qwen-VL with a randomly initialized action policy, while the other
starts from our UMI-pretrained checkpoint; the data split, action representation,
normalization, optimization schedule, and deployment protocol are otherwise
identical. Initialization is therefore the only controlled difference.
OpenPI-$\pi_{0.5}$ is included as a cross-architecture reference rather than as
part of this controlled comparison.

\begin{figure}[!t]
  \centering
  \includegraphics[width=0.98\linewidth]{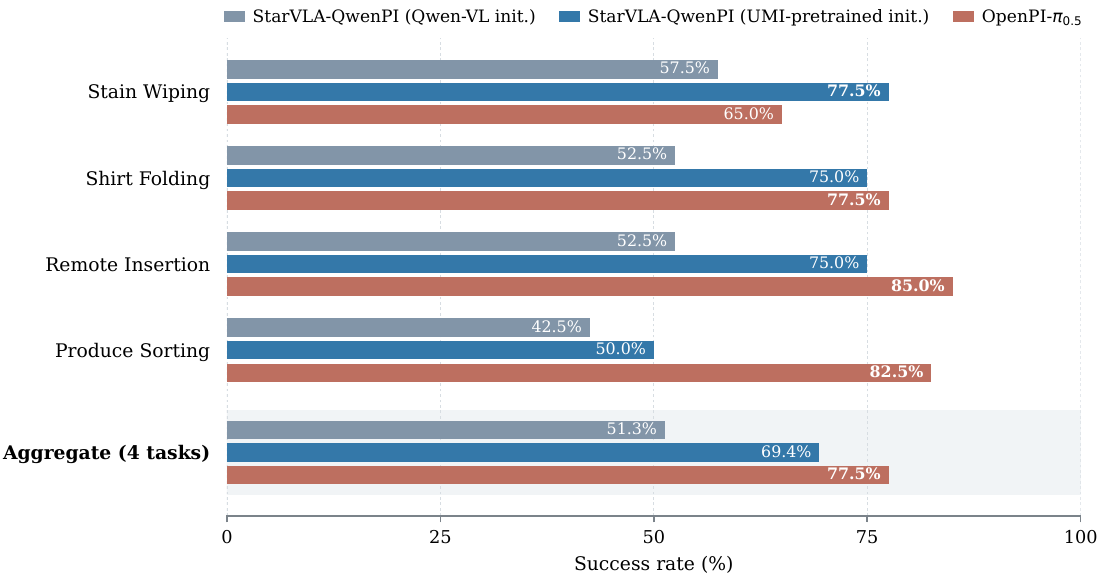}
  \caption{Real-robot success after UMI-only post-training. StarVLA-QwenPI is
  initialized either from Qwen-VL or our UMI-pretrained checkpoint;
  OpenPI-$\pi_{0.5}$ provides a cross-architecture reference. Each task contains
  40 rollouts per policy. Bold labels indicate the highest success rate within
  each group.}
  \label{fig:pretrained_downstream_success}
\end{figure}

Figure~\ref{fig:pretrained_downstream_success} shows a positive transfer effect on every
benchmark task. UMI pre-training raises aggregate StarVLA-QwenPI success by 18.1
percentage points, with particularly strong gains on wiping, folding, and
insertion. Because the architecture and post-training recipe are fixed, this
improvement can be attributed to the visual-motor initialization rather than
additional task-specific supervision. OpenPI-$\pi_{0.5}$ provides context for
the resulting performance level, but is not a controlled baseline.

The three evaluations connect pre-training exposure to deployment: prediction
improves throughout the training pass, the same trend extends to unseen tasks,
and the resulting checkpoint learns downstream tasks with less data. More
importantly, the task-family analysis identifies why transfer differs across
skills. Scaling the corpus is effective when it expands coverage of the
interaction dynamics that the policy must later reuse.

\FloatBarrier
\section{Discussion}
\label{sec:discussion}

\subsection*{Empirical Findings and Implications}
Across the four evaluated tasks and three tested backbones---two VLAs and one
WAM---HiFi-UMI serves as the sole task-specific post-training source and
achieves approximate aggregate parity with in-domain teleoperation. This
finding concerns the source of task-specific post-training data rather than the
complete pre-training history of the models: OpenPI-$\pi_{0.5}$ and LingBot-VA
retain their publicly released pre-trained initializations, while no
target-task teleoperation data are introduced in the HiFi-UMI post-training
conditions. Our additional HiFi-UMI pre-training is not required for this
aggregate post-training result; separately, on StarVLA-QwenPI, pre-training on
$4{,}000$ hours of HiFi-UMI further improves post-training data efficiency and
final deployment performance, shifting the efficiency--performance frontier
upward. These results also suggest that, once task-specific demonstrations are
sufficiently action-aligned, data composition and coverage may become as
important as volume alone. Characterizing which interaction dynamics a
deployable policy most depends on, and when that coverage saturates, remains a
central open question.

\subsection*{Limitations and Future Work}

\paragraph{Evaluation scope and generality.}
Our zero-robot post-training evidence covers four tabletop bimanual tasks and
three backbones under scene-level distribution shift; its generality to other
tasks, embodiments, and shifts remains untested. The pre-training evidence is
narrower: scaling and downstream gains are measured only on StarVLA-QwenPI.
Broader tasks and embodiments, together with additional VLA and WAM backbones,
are needed to establish the scope of both findings.

\paragraph{Task-level statistical resolution.}
Each task--policy pair has $40$ rollouts, so one additional success changes the
estimated rate by $2.5$ percentage points. The resulting uncertainty limits
fine-grained task-level comparisons, because a few outcomes can reverse the
ordering of methods separated by small gaps. We therefore treat per-task
differences as descriptive and base the parity claim on aggregate evidence
across tasks. More trials per condition would narrow the uncertainty and allow
task-level effects to be characterized more precisely.

\paragraph{Fidelity is validated as a whole, not decomposed.}
We treat fidelity as a design principle realized jointly by trajectory accuracy,
inter-gripper relative pose, synchronization, and field of view, and we do not
isolate these factors through controlled degradation. Our results thus show
\emph{that} high fidelity suffices, but not \emph{how much} of each property a
deployable policy requires. A systematic ablation---selectively degrading each
factor while holding sample count and scene coverage fixed---would quantify its
marginal contribution and turn ``high fidelity helps'' into an actionable
specification of the fidelity required for deployment.

\paragraph{Data efficiency and transfer across post-training sources.}
Our parity comparison is not sample-matched: without pre-training, UMI-only
post-training uses roughly ten times as many demonstrations as the teleoperation
baseline, so the result compares practical data-production pipelines rather than
per-trajectory efficiency. Large-scale UMI pre-training sharply improves both
efficiency and attainable performance, but two questions remain: whether
the gains continue as the pre-training corpus grows or eventually saturate, and
whether the same initialization benefits post-training on real-robot
teleoperation data as much as it benefits post-training on UMI data. Testing both
at larger scales would distinguish a generally reusable initialization effect
from a benefit specific to matched-domain training.

\section{Conclusion}
\label{sec:conclusion}

We revisit the assumption that robot-free demonstrations can seed but not finish
a policy without a teleoperated post-training anchor. We argue that the limitation
is \emph{fidelity}, not the robot-free setting. HiFi-UMI tests this claim with
portable capture co-designed for trajectory accuracy, inter-gripper relative pose,
synchronization, and field of view; its pipeline has generated over $20{,}000$
hours.

Across three VLA and WAM backbones, policies post-trained \emph{solely} on
HiFi-UMI match teleoperation within roughly $3$ percentage points, with
differences of both signs within sampling noise, and reach $85\%$
on precision insertion despite scene shift and zero teleoperated data. Separately,
$4{,}000$ hours of pre-training cut action error on ten unseen tasks by $41\%$
and raise real-robot success by $18.1$ percentage points at matched post-training data,
reaching the scratch-initialized baseline with one quarter of the task data.
Thus, robot-free data of this fidelity can support deployable manipulation,
not just pre-training. We release HiFi-UMI-2K, a $2{,}000$-hour,
microsecond-synchronized, replayable, ultra-wide-FoV subset of that corpus.

\clearpage
\phantomsection
\label{sec:contributions}
\section*{Author Contributions}
\addcontentsline{toc}{section}{Author Contributions}

\noindent\textbf{Core Contributors}\\
Yuteng Wei\textsuperscript{*}, Jinming Ma\textsuperscript{*},
Jiawei Wang\textsuperscript{*\textdagger}, Weitao Zhou\textsuperscript{*\textdagger},
Yushen Zuo, Ke Rui, Minglei Li\textsuperscript{\textdagger\,\ding{41}}.

\medskip
\noindent\textbf{Contributors}\\
Jinhao Zhang, Zhikang Pan, Xiang Wang, Haoran Jia, Huan Du, Zicheng Zeng,
Jun Ma, Guiyu Qin, Di Zhang, Xiaofei Li.

\bigskip
\noindent\textsuperscript{*}Equal contribution.\quad
\textsuperscript{\textdagger}Project leaders.\quad
\textsuperscript{\ding{41}}Corresponding author.

\medskip
\noindent\textit{Correspondence:} \email{liminglei@simpleai.tech}.

\bibliography{references}
\bibliographystyle{bibstyle}

\end{document}